\documentclass{article}
\usepackage[preprint]{neurips_2026}

\usepackage[utf8]{inputenc} % allow utf-8 input
\usepackage[T1]{fontenc}    % use 8-bit T1 fonts
\usepackage{hyperref}       % hyperlinks
\usepackage{url}            % simple URL typesetting
\usepackage{booktabs}       % professional-quality tables
\usepackage{amsfonts}       % blackboard math symbols
\usepackage{nicefrac}       % compact symbols for 1/2, etc.
\usepackage{microtype}      % microtypography
\usepackage{xcolor}         % colors
\usepackage{amsmath}
\usepackage{graphicx} 
\title{Assessing and Mitigating Miscalibration in LLM-Based Social Science Measurement}
\author{%
  Jinyuan Wang$^\ast$ \\
  The Hong Kong University of Science and Technology \\
  \texttt{jinyuanwang@ust.hk} \\
  \And
  Ningyuan Deng$^\ast$ \\
  The Hong Kong University of Science and Technology \\
  \texttt{ningyuandeng@ust.hk} \\
  \And
  Yi Yang \\
  The Hong Kong University of Science and Technology \\
  \texttt{imyiyang@ust.hk} \\
}

\begin{document}

\maketitle

\begin{abstract}

Large language models (LLMs) are increasingly used in social science as scalable measurement tools for converting unstructured text into variables that can enter standard empirical designs. Measurement validity demands more than high average accuracy, which requires well calibrated confidence that faithfully reflects the empirical probability of each measurement being correct. This paper studies the model miscalibration in LLM-based social science measurement. We begin with a case study on FOMC and show that confidence based filtering can change downstream regression estimates when LLM confidence is miscalibrated. We then audit calibration across 14 social science constructs covering both proprietary models, including GPT-5-mini, DeepSeek-V3.2, and open source models. Across tasks and model families, reported confidence is poorly aligned with tolerance-based correctness. As a simple mitigation, we propose a soft label distillation pipeline for calibrating Bert with LLM. The method converts an LLM score and its verbalized confidence into a soft target distribution, then trains a smaller discriminative classifier on encoder models for these targets. Averaged across datasets, this approach reduces ECE by 43.2\% and Brier by 34.0\%. These results suggest that LLM-based social science pipelines should treat calibration as part of measurement validity, rather than as an optional post-processing concern.

\end{abstract}

\section{Introduction}

Large language models (LLMs) are increasingly used as measurement tools in social science where researchers prompt models to convert unstructured text (and other qualitative inputs) into continuous numeric constructs that can enter standard empirical designs \citep{asirvatham2026gpt, bail2024can}. Recent evidence suggests that LLM automated labels can approach expert annotations in specific settings, such as FOMC sentence classification and psychological constructs (sentiment, discrete emotions, offensiveness, and moral foundations) \citep{rathje2024gpt,hansen2023fedspeak}. However human like scoring alone does not guarantee measurement validity. If the LLM's confidence is miscalibrated, then high confidence subsets can introduce systematic bias into downstream regression estimates rather than improve them. This requirement goes beyond aggregate accuracy and demands that the model's reported confidence be strictly calibrated.

% Calibration has been studied extensively in general NLP and QA settings, including verbalized confidence elicitation \citep{kadavath2022language,tian2023just}, agreement and prompt sensitive calibration \citep{xia2025influences}, fidelity aware confidence estimation \citep{zhang2024ufcalibration}, and post hoc calibration tuning \citep{kapoor2024calibrationtuning,manggala2025qacalibration}. 
% % Rank-based uncertainty assessment and activation-level calibration further show that confidence quality depends strongly on how uncertainty is represented \citep{huang2024rankcalibration,liu2024actcab}. 
% Recent work shows persistent calibration failures in long form QA and multi answer settings, even when overall answer quality improves \citep{muller2026benchmarking,wang2026mace,liu2026cure,xie2024calibrating}. Despite these technical advances, the downstream consequences of miscalibration in social science measurement remain critically underexplored. To the best of our knowledge, this is the first study to measure LLM calibration in the social science domain. When an LLM reports 90$\%$ confidence but is correct only 30$\%$, miscalibration is severe. Applying high confidence filtering in such cases does not improve estimation; instead, it introduces attenuation bias, shrinking regression coefficients toward zero and distorting policy estimation \citep{kapoor2024calibrationtuning,manggala2025qacalibration}.
While LLM calibration has been extensively studied in general NLP and QA \citep{kadavath2022language,tian2023just,xia2025influences,zhang2024ufcalibration,kapoor2024calibrationtuning,manggala2025qacalibration}, its downstream consequences in social science measurement remain critically underexplored. To our knowledge, this is among the first systematic calibration audits focused on LLM-based social science measurement. This gap matters because severe miscalibration can make naive high-confidence filtering unreliable. Instead, it introduces attenuation bias, artificially shrinking regression coefficients and distorting downstream policy estimation \citep{kapoor2024calibrationtuning,manggala2025qacalibration}.

In an FOMC sentence level hawkish vs dovish task, we show that filtering by different confidence alters the estimated relationship between the derived policy stance and CPI, producing materially different coefficients \citep{hansen2023fedspeak}. This example motivates our central question, can LLM confidence be reliably used as an uncertainty measure in social science pipelines?

To answer this question, we conduct a multi dataset empirical study on social science tasks with human labels and evaluate calibration under tolerance based correctness criteria. We focus on settings where outputs are used as measurements rather than pure QA answers \citep{asirvatham2026gpt,bail2024can}. Our results show a consistent mismatch between reported confidence and realized correctness across models and tasks, indicating that calibration failure is a practical threat to empirical validity in social science applications \citep{de2025study,kapoor2024calibrationtuning}.

Beyond diagnosis, we propose a simple mitigation pipeline. We first ask the LLM to provide both a score and self verbalized confidence. We then convert this pair into soft labels over discrete score classes, and train a BERT classifier on these soft targets. Intuitively, this distillation step separates semantic scoring from uncertainty representation and yields more stable calibration behavior than directly using prompted LLM predictions \footnote{Our dataset and codes are shown in \url{https://anonymous.4open.science/r/dataset_for_miscalibration-7849}}.

Our core contributions are threefold:
\begin{itemize}
    \item To the best of our knowledge, we are the first to formalize calibration as a critical reliability bottleneck in LLM-based social science measurement. We empirically demonstrate that naive confidence based filtering can severely distort downstream regression estimates.
    \item We provide a comprehensive calibration audit across 14 diverse social science constructs. This evaluation reveals a pervasive alignment calibration gap and exposes the resolution collapse inherent in standard post-hoc calibration methods.
    \item We compare different post-hoc calibration methods and introduce a lightweight, confidence based soft label distillation approach that distills knowledge from an LLM to a BERT model, improving calibration quality while preserving downstream task performance.
\end{itemize}

\section{Case Study for Miscalibration Distorts Regression Estimation in LLM-Based Social Science}

% To illustrate how miscalibration distorts regression estimation, we conduct a case study in which an LLM answers questions and simulates a social science regression task. Research directly uses an LLM prompt to measure the stance of policy text~\citep{hansen2023fedspeak}. Our key message is simple: when confidence is miscalibrated, confidence-based sample selection can materially change downstream regression coefficients. Calibration error is not only a prediction-quality issue, it is an inference-validity issue.

To demonstrate how miscalibration distorts regression estimation, we present a case study in which an LLM answers questions for a social science regression task. Our key finding is that when confidence is miscalibrated, confidence based sample selection can materially alter downstream regression coefficients. Thus, calibration error is not merely a prediction quality issue, it is an estimation issue. While prior research directly uses LLM prompts to measure the stance of policy text~\citep{hansen2023fedspeak}, our work reveals the critical risk for downstream estimation.

This analysis uses CPI change as the independent variable to predict Federal Open Market Committee (FOMC) monetary policy stance scores (0 = dovish, 100 = hawkish). We use GPT-5-nano to generate LLM results. To examine how calibration affects downstream regression, we filter sentences by confidence thresholds above 90, aggregate them to the daily level, and run OLS: $\text{Stance} \sim \text{CPI change}$, where Stance is the dependent variable and CPI change is the independent variable. Following \citep{Shah2023TrillionDW}, we compute the daily stance score as $(\text{\ count of hawkish sentences} - \text{count of dovish sentences})/ \text{total sentences}$, where hawkish sentences are those with stance $> 50$  and dovish sentences are those with stance $< 50$. Ground truth is estimated using human annotated labels from the same FOMC dataset without confidence filtering. The dataset covers the period from 2018 to 2020 and comprises approximately 4,000 sentences taken from 37 files of the FOMC meeting minutes.

\begin{table}[htbp]
\centering
\begin{tabular}{lcccccc}
\toprule
\textbf{Method} & \textbf{$\beta$ (Coefficient)} & \textbf{t-statistic}& \textbf{R²} \\
\midrule
\textbf{ Ground Truth}& 0.0855 & 9.0631  & 0.7012 \\
\midrule
GPT-5-nano& 0.0477 & 7.8213 &0.6361 \\
% Confidence Threshold 10 & 37 & 0.0487*** & 7.8728 & 0.0000 & 0.6391 \\
% Confidence Threshold 20 & 37 & 0.0486*** & 7.8511 & 0.0000 & 0.6378 \\
% Confidence Threshold 30 & 37 & 0.0486*** & 7.8519 & 0.0000 & 0.6379 \\
% Confidence Threshold 40 & 37 & 0.0486*** & 7.8519 & 0.0000 & 0.6379 \\
% Confidence Threshold 50 & 37 & 0.0488*** & 7.8799 & 0.0000 & 0.6395 \\
% Confidence Threshold 60 & 37 & 0.0490*** & 7.8759 & 0.0000 & 0.6393 \\
% Confidence Threshold 70 & 37 & 0.0485*** & 8.0215 & 0.0000 & 0.6477 \\
% GPT-5-nano Confidence Threshold 80  & 0.0396 & 7.7968 & 0.0000 & 0.6346 \\
GPT-5-nano w/ confidence filter & 0.0261 & 4.7935 & 0.3963 \\
\bottomrule
\end{tabular}
\caption{The regression results show the change in the CPI against the FOMC hawk-dove stance. The GPT-5-nano used all datasets, w/ confidence means filtering data with a threshold of 90.}
\label{tab:regression_results}
\end{table}

% Across all thresholds from $\geq 0$ to $\geq 80$ and $\geq 90$, the coefficient estimates remain substantially below the Ground Truth, 0.0855. While the coefficient is relatively stable between $\geq 0$ and $\geq 70$ (around 0.048), it declines to 0.0396 at $\geq 80$ and collapses to 0.0261 at $\geq 90$, where R² drops sharply from 0.64 to 0.396. These results indicate that the LLM systematically underestimates the relationship between CPI changes and FOMC stance across all confidence levels, and aggressive confidence filtering ($\geq 90$) severely degrades both coefficient magnitude and model fit.

% ~\citep{neuhaus1999bias}
% We filter model outputs by confidence before downstream analysis. A miscalibrated model can introduce selection bias and distort causal estimates: "more confident" no longer means "more correct," so retained samples may be systematically biased. A practical consequence is that analysts may trust a high-confidence subset that still carries large measurement error, analogous to a 90$\%$ confidence$30\%$ accuracy failure pattern. In our case, the threshold sweep shows exactly why this matters: coefficient estimates vary materially with confidence filtering, and aggressive filtering ($\geq 90$) can collapse sample size and destabilize inference. This highlights that calibration is an essential complement to traditional regression measures, especially in policy-relevant settings where unbiased estimation matters.

% \paragraph{Formal Definition}

The estimated slope from the unfiltered GPT-5-nano model  (\(\beta=0.0477\)) is below the ground truth coefficient  (\(\beta=0.0855\)). Applying a 90 confidence filter further attenuates the estimate, reducing the coefficient to 0.0261. Model fit deteriorates correspondingly \(R^2\) drops from 0.7012 for ground truth to 0.6361 for all datasets and further to 0.3963 for w/ confidence filter. Importantly, the t-statistic remains high for the unfiltered regression model with 7.82 and lower for the filtered regression model with 4.79, indicating that the issue is not statistical insignificance but effect size attenuation. In other words, the relationship remains detectable, but its magnitude is systematically pushed toward zero as the confidence filter is applied.

This pattern is consistent with attenuation bias in the outcome mismeasurement problem \citep{neuhaus1999bias}, which means that estimated associations are biased and commonly attenuated toward zero. In our social science case, confidence filtering is intended to improve label quality, but with miscalibrated confidence, it can instead induce selective measurement error. As a result, filtering by confidence can mechanically weaken estimated relationships, exactly as observed in the monotonic coefficient shrinkage for 90 confidence filter. This implies that social science studies that directly threshold LLM confidence may report attenuated effects even when the true relationship is stronger.

\section{Definitions}

The problem we address in this paper is the calibration of continuous numeric social science measurements generated by LLMs. For a given input $x$, let $y \in \mathbb{R}$ be the ground truth of the social science measurement annotated by humans, and $\hat{y} = f(x)$ be the prediction score given by the LLM. We denote $conf \in [0, 1]$ as the model's confidence estimate that the prediction result falls within the error tolerance, where $\epsilon > 0$ is a defined tolerance threshold (e.g., 10 points on a 0-100 scale) that can be adjusted according to the practical needs of social science researchers. A measurement tool is considered perfectly calibrated under tolerance $\epsilon$ if, for a confidence level $q \in [0, 1]$ given by the LLM, its actual accuracy (i.e., the probability that the absolute difference between the prediction and the ground truth is less than or equal to $\epsilon$) is exactly equal to $q$:
\begin{equation}
    \mathbb{P}(|\hat{y} - y| \le \epsilon \mid conf = q) = q
\end{equation}

\paragraph{Tolerance based Expected Calibration Error ($\text{T-ECE}_\epsilon$).} 
To empirically quantify the deviation from perfect calibration, we adapt the standard Expected Calibration Error (ECE) to accommodate continuous measurements with error boundaries. Given $n$ predictions, we partition the samples into $M$ equally spaced confidence bins $B_1, \dots, B_M$. The Tolerance based ECE ($\text{T-ECE}_\epsilon$) is then defined as the expected absolute difference between the empirical tolerance accuracy and the average confidence within each bin:
\begin{equation}
    \text{T-ECE}_\epsilon = \sum_{m=1}^{M} \frac{|B_m|}{n} \left| \text{acc}_\epsilon(B_m) - \text{conf}(B_m) \right|,
\end{equation}
where $\text{conf}(B_m)$ is the average confidence score of the samples in bin $B_m$, and the empirical tolerance accuracy $\text{acc}_\epsilon(B_m)$ is calculated as the proportion of predictions within the bin that fall inside the tolerance threshold $\epsilon$:
\begin{equation}
    \text{acc}_\epsilon(B_m) = \frac{1}{|B_m|} \sum_{i \in B_m} \mathbb{I}(|\hat{y}_i - y_i| \le \epsilon).
\end{equation}

\paragraph{Brier Score.} 
To assess the overall predictive accuracy and calibration of the confidence estimates, we employ the Brier Score, which measures the mean squared error between the predicted confidence and the actual binary outcome of the tolerance-based correctness. For a set of $n$ predictions, it is defined as:
\begin{equation}
    \text{Brier} = \frac{1}{n} \sum_{i=1}^{n} (conf_i - o_i)^2
\end{equation}
where $o_i = \mathbb{I}(|\hat{y}_i - y_i| \le \epsilon)$ is the binary indicator of whether the $i$-th prediction falls within the tolerance $\epsilon$. A lower Brier Score indicates better-calibrated and more accurate probability estimates.

\paragraph{Machine Human Correlation (MH).} 
% Following \citep{asirvatham2026gpt}, we measure the alignment between the LLM’s predicted scores and human-annotated ground truth using the Machine-Human Correlation (MH). This metric is calculated as the Spearman correlation coefficient between $\hat{y}$ and $y$ across all samples:
% \begin{equation}
%     \text{MH} = \frac{\sum_{i=1}^{n} (\hat{y}_i - \bar{\hat{y}})(y_i - \bar{y})}{\sqrt{\sum_{i=1}^{n} (\hat{y}_i - \bar{\hat{y}})^2 \sum_{i=1}^{n} (y_i - \bar{y})^2}}
% \end{equation}
% While MH captures the linear consistency of the measurement tool, it does not account for the absolute calibration of the scores, making it a necessary but insufficient condition for reliable social science measurement.
Following \citep{asirvatham2026gpt}, we measure the alignment between the LLM's predicted scores and human annotated ground truth using the Machine Human Correlation (MH). This metric is calculated as the Spearman rank correlation coefficient between $\hat{y}$ and $y$ across all samples:
\begin{equation}
\text{MH} = 1 - \frac{6 \sum_{i=1}^{n} d_i^2}{n(n^2 - 1)}
\end{equation}
where $d_i = \text{rg}(\hat{y}_i) - \text{rg}(y_i)$ is the difference between the ranks of the predicted score and the ground truth for the $i$-th sample, and $n$ is the total number of samples.

\section{Observing LLM-based Measurement Miscalibration}

\subsection{Social Science Measurement Calibration Dataset}

To systematically evaluate LLM calibration, we introduce the \textbf{Social Science Measurement Calibration Dataset (SSMCD)}, which spans 14 constructs across 9 source datasets from~\citep{PavlickAndTetreault-2016:TACL, danescu2013computational, buechel2017emobank, hossain2019president, kennedy2020constructing, ashida2022towards, DBLP:journals/corr/abs-1911-11408, hansen2023fedspeak, le2023uncovering} (see Table~\ref{tab:dataset_summary} for details). 
All instances are grounded in human annotations. We sample $N=1{,}000$ instances per dataset to ensure distributional stability, retaining 85\% of the data for corpora smaller than 2,000. 
Continuous construct scores are linearly rescaled to $[0, 100]$. Additionally, to mitigate computational costs associated with subsequent experimental designs, we extract 100 sample stratified subsets from four representative datasets (\textit{EmoBank}, \textit{CHASM}, \textit{Pavlick Formality}, and \textit{Stanford Politeness}).
Appendix~\ref{sec:constructs_definition} provides detailed dataset information and construct definitions.

\subsection{Models}

We evaluate a diverse suite of LLMs across two primary access paradigms to test the generalizability of calibration deficits: API based models, including GPT-5-nano (GPT-5-nano), GPT-5-mini (GPT-5-mini), DeepSeek-V3.2 (DeepSeek-V3.2), and Qwen-3.5-Flash (Qwen3.5-Flash), and locally deployed open source models, including Qwen-2.5-7B-Instruct (Qwen2.5-7B), Ministral-8B-Instruct (Ministral-8B), DeepSeek-R1-Distill-Qwen-7B (DeepSeek-R1-7B), and Gemma-2-9B-IT (Gemma-2-9B).

\subsection{Measurement Method}

Unlike standard probabilistic classifiers, LLMs do not natively output calibrated confidence scores for continuous measurements. Therefore, we consider four approaches to construct confidence proxies that estimate the reliability of LLM generated measurements:

\paragraph{Verbalized Confidence (Verbal).} We directly elicit a confidence score from the LLM by appending an additional accuracy estimation construct to the evaluation prompt. As detailed in Appendix~\ref{sec:prompt_template}, this instructs the model to estimate the probability on a 0--100 scale that its predicted score falls within the specified error tolerance~\citep{tian2023just, asirvatham2026gpt}.

\paragraph{Resampling Confidence (Resampling).} We measure consistency across multiple stochastic forward passes~\citep{xiong2023can}. For continuous outputs, we apply a sliding window of width $2\epsilon$ across the sampled scores to locate the region of maximum density. Confidence is defined as the proportion of samples within this maximal window, and the final measurement is their mean.

% \paragraph{Logit based Confidence (Logit (Geom.)).} Following \citep{liu2023litcab}, we quantify confidence as the geometric mean of the generation probabilities across all $L$ tokens in the predicted score:
% \begin{equation}
%     p(y|x) = \sqrt[L]{\prod_{t=1}^{L} p(y_t | x, y_{<t})}
% \end{equation}
% This probability derived metric reflects the model's implicit uncertainty grounded in its next token distribution. It serves as a continuous and computationally efficient measure that avoids the overhead of additional model queries.
\paragraph{Logit-based Confidence (Logit (Geom.))} Following \citep{liu2023litcab}, we compute confidence as the geometric mean of the generation probabilities across all tokens in the predicted score.

\paragraph{Logit-based Confidence (Logit (P-true)).} Following \citep{kadavath2022language}, we prompt the model to evaluate the correctness of its own generated measurement. The confidence is then calculated as the softmax-normalized probability of the ``True'' token relative to the ``False'' token from the model's output logits.

\subsection{Main Results}
Table \ref{tab:main_results} presents the macro-averaged measurement and calibration performance across all 14 social science constructs. Our empirical audit of these confidence proxies yields primary findings regarding the current state of LLM calibration.

\begin{table}[htbp]
\centering
\caption{Macro-averaged calibration and measurement performance across 14 social science constructs. The evaluation contrasts verbalized confidence, resampling variance, and logit-based probabilities (\textit{Geom.} and \textit{P-true}) across both proprietary APIs and locally deployed open source models. Lower $T\text{-ECE}_\epsilon$ and Brier scores indicate superior calibration. Bold values denote the best performance.}
\label{tab:main_results}
\resizebox{\textwidth}{!}{%
\begin{tabular}{@{}llcccccc@{}}
\toprule
 &  & \multicolumn{3}{c}{\textbf{Full Dataset}} & \multicolumn{3}{c}{\textbf{Sample Dataset}} \\
\cmidrule(lr){3-5} \cmidrule(l){6-8}
\textbf{Model} & \textbf{Method} & \textbf{$\text{T-ECE}_\epsilon \downarrow$} & \textbf{Brier $\downarrow$} & \textbf{MH $\uparrow$} & \textbf{$\text{T-ECE}_\epsilon \downarrow$} & \textbf{Brier $\downarrow$} & \textbf{MH $\uparrow$} \\ 
\midrule
GPT-5-nano             & Verbal           & 0.342 & 0.351 & 0.501 & 0.354 & 0.359 & 0.451 \\
                       & Self-Random      &   -   &   -   &   -   & 0.568 & 0.558 & 0.478 \\
\addlinespace
GPT-5-mini             & Verbal           & 0.466 & 0.440 & \textbf{0.605} & 0.466 & 0.438 & \textbf{0.574} \\
                       & Self-Random      &   -   &   -   &   -   & 0.654 & 0.648 & 0.559 \\
\addlinespace
DeepSeek-V3.2          & Verbal           & 0.396 & 0.400 & 0.537 & 0.366 & 0.382 & 0.467 \\
                       & Self-Random      &   -   &   -   &   -   & 0.431 & 0.413 & 0.501 \\
\addlinespace
Qwen3.5-Flash          & Verbal           & 0.516 & 0.495 & 0.587 & 0.528 & 0.503 & 0.544 \\
                       & Self-Random      &   -   &   -   &   -   & 0.604 & 0.588 & 0.548 \\
\addlinespace
Qwen2.5-7B             & Verbal           & 0.459 & 0.441 & 0.418 & 0.417 & 0.411 & 0.402 \\
                       & Self-Random      &   -   &   -   &   -   & 0.426 & 0.414 & 0.455 \\
                       & Logit (Geom.)    & \textbf{0.180} & 0.262 & 0.427 & \textbf{0.186} & 0.265 & 0.409 \\
                       & Logit (P-true)   & 0.597 & 0.580 & 0.427 & 0.626 & 0.603 & 0.409 \\
\addlinespace
Ministral-8B           & Verbal           & 0.540 & 0.508 & 0.485 & 0.511 & 0.495 & 0.465 \\
                       & Self-Random      &   -   &   -   &   -   & 0.455 & 0.383 & 0.291 \\
                       & Logit (Geom.)    & 0.228 & 0.261 & 0.254 & 0.230 & 0.254 & 0.254 \\
                       & Logit (P-true)   & 0.357 & 0.321 & 0.254 & 0.374 & 0.327 & 0.254 \\
\addlinespace
DeepSeek-R1-7B         & Verbal           & 0.425 & 0.412 & 0.287 & 0.378 & 0.367 & 0.283 \\
                       & Self-Random      &   -   &   -   &   -   & 0.538 & 0.479 & 0.236 \\
                       & Logit (Geom.)    & 0.207 & \textbf{0.229} & 0.262 & 0.234 & \textbf{0.231} & 0.260 \\
                       & Logit (P-true)   & 0.693 & 0.665 & 0.262 & 0.723 & 0.691 & 0.260 \\
\addlinespace
Gemma-2-9B             & Verbal           & 0.478 & 0.453 & 0.424 & 0.498 & 0.476 & 0.418 \\
                       & Self-Random      &   -   &   -   &   -   & 0.565 & 0.543 & 0.443 \\
                       & Logit (Geom.)    & 0.212 & 0.258 & 0.425 & 0.216 & 0.256 & 0.429 \\
                       & Logit (P-true)   & 0.513 & 0.482 & 0.425 & 0.495 & 0.454 & 0.429 \\
\bottomrule
\end{tabular}%
}
\end{table}

\begin{figure}[htbp]
    \centering
    \includegraphics[width=1\linewidth]{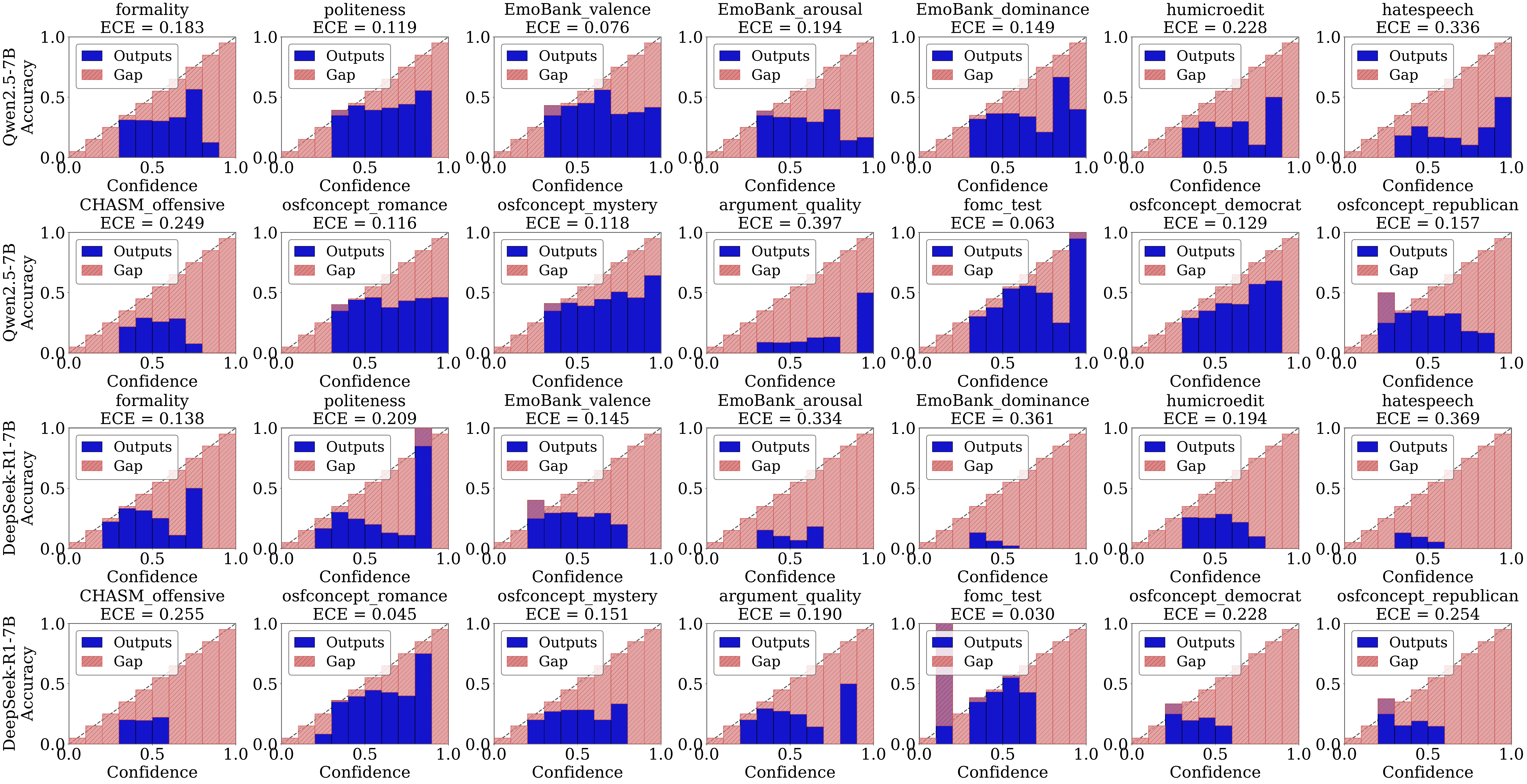}
    \caption{Reliability diagrams for Qwen-2.5-7B-Instruct and DeepSeek-R1-Distill-Qwen-7B using the Logit (Geom.) proxy across 14 constructs. While these configurations yield the lowest $T\text{-ECE}_\epsilon$ and Brier scores respectively (see Table~\ref{tab:main_results}), their predictions deviate from the ideal diagonal.}
    \label{fig:qwen_deepseek_logit_geometric_mean_4x7}
\end{figure}

\textbf{Widespread Miscalibration.} Across the eight models and all evaluated confidence proxies, we observe a systemic lack of sufficient calibration. While logit based approaches numerically outperform verbalized confidence with Qwen-2.5-7B-Instruct (Logit Geom.), achieving the lowest $T\text{-ECE}_\epsilon$ of 0.180, and DeepSeek-R1-7B (Logit Geom.) attaining the lowest Brier score of 0.229. As visually confirmed by the reliability diagrams for these top performing configurations (Figure~\ref{fig:qwen_deepseek_logit_geometric_mean_4x7}), the predictions consistently deviate from the ideal diagonal line, exhibiting severe miscalibration. This widespread deficit confirms that whether elicited via natural language or extracted from internal logits, the raw confidence estimates of current LLMs remain highly unreliable for rigorous downstream sample filtering.

\textbf{The Alignment-Calibration Gap.} We observe a stark dissociation between predictive alignment (MH) and absolute calibration ($T\text{-ECE}_\epsilon$, Brier score). While proprietary models like GPT-5-mini achieve state-of-the-art rank ordering (highest MH with 0.605), this relative alignment fails to yield reliable probabilities, as evidenced by severe miscalibration ($T\text{-ECE}_\epsilon$ with 0.466, Brier with 0.438). Conversely, smaller open source models using logit methods achieve vastly superior calibration despite weaker alignment (e.g., Qwen-2.5-7B-Instruct attains a $T\text{-ECE}_\epsilon$ of 0.180 but an MH of only 0.427). This implies that a model's measurement capability is not a reliable proxy for its confidence or trustworthiness.

% \begin{figure}
%     \centering
%     \includegraphics[width=1\linewidth]{figures/qwen_deepseek_logit_geometric_mean_4x7.pdf}
%     \caption{Reliability diagram of Qwen-2.5-7B-Inst. across 14 constructs. Despite achieving the state-of-the-art $T\text{-ECE}_\epsilon$ ($0.180$) in our benchmarking (see Table~\ref{tab:main_results}), the model still exhibits substantial calibration misalignments on the full dataset. The consistent deviation from the ideal diagonal line reveals that even top-performing open source models suffer from systemic over or under confidence when quantifying complex social science constructs.}
%     \label{fig:qwen_deepseek_logit_geometric_mean_4x7}
% \end{figure}

\subsection{Post-hoc Calibration}

\textbf{Methodology and Calibration Setup.} To determine whether calibration deficits are structural or scaling related, we evaluate four post-hoc methods: Platt scaling~\citep{platt1999probabilistic}, Beta calibration~\citep{kull2017beta}, Isotonic regression~\citep{barlow1972isotonic}, and Temperature scaling~\citep{ding2021local}. As shown in Table~\ref{tab:calibration_results}, each calibrator is trained on a dedicated 1,000 sample stratified subset per construct. Independent parameter fitting (\textit{e.g.}, $A, B, T$) is essential to ensure that scaling functions are tailored to the representative distribution.

% \textbf{Synthesized Results and the Resolution Collapse.} As shown in Table~\ref{tab:calibration_results}, post-hoc methods—especially Platt and Beta scaling—dramatically reduce $T\text{-ECE}_\epsilon$, often from $>0.400$ to $<0.030$. However, this low error masks a pervasive \textit{ECE Collapse} (Figure~\ref{fig:reliability_posthoc}). With the exception of Temperature scaling, these calibrators typically collapse diverse model outputs into a single bin near the dataset's base rate~\citep{kumar2019verified}. While this mathematically minimizes ECE, it destroys the model's discriminative resolution, yielding "safe" but uninformative predictions. 

\begin{figure}
    \centering
    \includegraphics[width=1\linewidth]{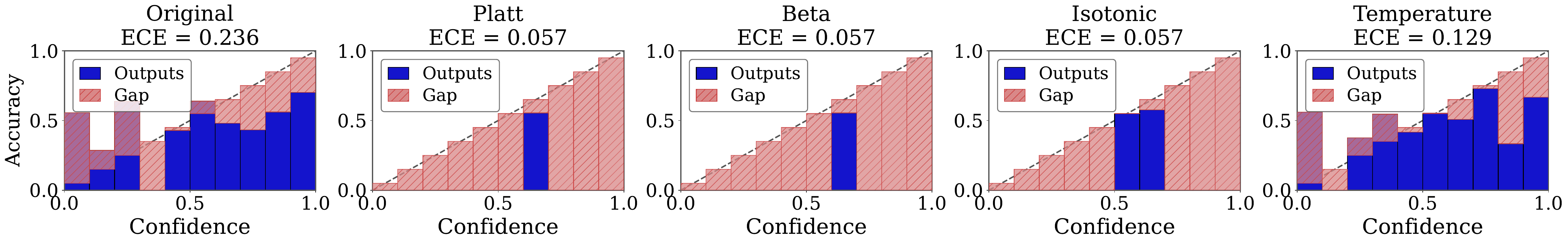}
    \caption{ Reliability diagrams for GPT-5-nano on the Formality task. The original verbalized confidence (left) shows significant under-confidence and non-linear scaling. While parametric methods like Platt and Beta (center) successfully align predictions with the ideal diagonal, they achieve this at the cost of near resolution collapse.}
  
    % Reliability diagrams of GPT-5-nano on \textit{Formality} before and after post-hoc calibration. The original verbalized confidence (left) exhibits significant under-confidence and non-linear scaling, while parametric methods like Platt and Beta (center) successfully align the predictions with the ideal diagonal but achieving near collapse.
    
     \label{fig:reliability_posthoc}
\end{figure}

% To expose this collapse, we evaluate calibration via both $T\text{-ECE}_\epsilon$ (alignment) and Brier Score (penalizing resolution loss). Although Platt and Beta scaling achieve the lowest $T\text{-ECE}_\epsilon$, their disproportionately high Brier scores reveal they merely game the ECE metric without improving intrinsic predictive value. Thus, reliable measurement requires minimizing $T\text{-ECE}_\epsilon$ without sacrificing resolution, a balance where rank-preserving approaches like Temperature scaling prove strictly superior.
\textbf{Synthesized Results and the Resolution Collapse.} As detailed in Table~\ref{tab:calibration_results}, post-hoc methods particularly Platt and Beta scaling drastically reduce $T\text{-ECE}_\epsilon$, often from above $0.400$ to below $0.030$. However, this sharp reduction frequently masks a pervasive Resolution Collapse (Figure~\ref{fig:reliability_posthoc}). With the notable exception of Temperature scaling, these calibrators tend to collapse diverse model outputs into a narrow range near the dataset's base rate~\citep{kumar2019verified}, which leads to confidence scores becoming similar for all predictions. While this mathematically minimizes ECE, it destroys the model's discriminative resolution, yielding right but uninformative confidence.

To expose this collapse, we evaluate calibration through the both $T\text{-ECE}_\epsilon$ (alignment) and Brier score (resolution). Although Platt and Beta scaling achieve the lowest $T\text{-ECE}_\epsilon$, their disproportionately high Brier scores reveal that they merely optimize for the ECE metric without improving intrinsic predictive utility. Reliable social science measurement requires a balance between error alignment and resolution. In this context, rank preserving approaches such as Temperature scaling prove strictly superior because they maintain the model's discriminative power while improving calibration.

\begin{table}[htbp]
\centering
\caption{Macro averaged calibration performance across 9 datasets after post-hoc alignment. Each calibration method is trained on a dedicated 1,000 sample stratified subset per dataset. Results demonstrate that parametric methods (Platt and Beta) consistently outperform Isotonic regression and Temperature scaling in reducing $T\text{-ECE}_\epsilon$.}
\label{tab:calibration_results}
\resizebox{\textwidth}{!}{%
\small
\begin{tabular}{@{}ll*{5}{cc}@{}}
\toprule
\textbf{Model} & \textbf{Method} & \multicolumn{2}{c}{\textbf{Original}} & \multicolumn{2}{c}{\textbf{Platt}} & \multicolumn{2}{c}{\textbf{Beta}} & \multicolumn{2}{c}{\textbf{Isotonic}} & \multicolumn{2}{c}{\textbf{Temperature}} \\
\cmidrule(lr){3-4} \cmidrule(lr){5-6} \cmidrule(lr){7-8} \cmidrule(lr){9-10} \cmidrule(l){11-12}
 & & \textbf{ECE} $\downarrow$ & \textbf{Brier} $\downarrow$ & \textbf{ECE} $\downarrow$ & \textbf{Brier} $\downarrow$ & \textbf{ECE} $\downarrow$ & \textbf{Brier} $\downarrow$ & \textbf{ECE} $\downarrow$ & \textbf{Brier} $\downarrow$ & \textbf{ECE} $\downarrow$ & \textbf{Brier} $\downarrow$ \\
\midrule
DeepSeek-R1-7B        & Logit (Geom.)   & 0.219 & \textbf{0.231} & 0.028 & 0.170 & 0.023 & 0.170 & 0.029 & 0.172 & \textbf{0.156} & \textbf{0.220} \\
                      & Logit (P-true)  & 0.695 & 0.663 & 0.024 & 0.170 & 0.029 & 0.168 & 0.029 & 0.168 & 0.274 & 0.252 \\
                      & Verbal          & 0.389 & 0.376 & 0.026 & 0.157 & 0.029 & 0.159 & 0.031 & 0.161 & 0.283 & 0.250 \\
\addlinespace
Qwen2.5-7B            & Logit (Geom.)   & \textbf{0.194} & 0.261 & 0.032 & 0.211 & 0.032 & 0.211 & 0.045 & 0.214 & 0.183 & 0.250 \\
                      & Logit (P-true)  & 0.622 & 0.599 & 0.040 & 0.214 & 0.038 & 0.213 & 0.047 & 0.215 & 0.199 & 0.253 \\
                      & Verbal          & 0.441 & 0.420 & 0.033 & 0.215 & 0.033 & 0.216 & 0.039 & 0.217 & 0.172 & 0.251 \\
\addlinespace
DeepSeek-V3.2         & Verbal          & 0.431 & 0.419 & 0.024 & 0.193 & 0.022 & 0.193 & 0.030 & 0.194 & 0.212 & 0.251 \\
\addlinespace
Gemma-2-9B            & Logit (Geom.)   & 0.250 & 0.256 & \textbf{0.012} & 0.171 & \textbf{0.011} & 0.172 & \textbf{0.020} & 0.173 & 0.252 & 0.248 \\
                      & Logit (P-true)  & 0.565 & 0.519 & 0.029 & 0.169 & 0.023 & 0.169 & \textbf{0.020} & 0.170 & 0.265 & 0.251 \\
                      & Verbal          & 0.502 & 0.463 & 0.017 & 0.167 & 0.022 & 0.169 & 0.029 & 0.170 & 0.261 & 0.247 \\
\addlinespace
GPT-5-mini            & Verbal          & 0.533 & 0.485 & 0.031 & 0.157 & 0.023 & 0.160 & 0.026 & 0.162 & 0.256 & 0.249 \\
GPT-5-nano            & Verbal          & 0.403 & 0.376 & 0.023 & 0.161 & 0.022 & 0.161 & 0.025 & 0.162 & 0.251 & 0.258 \\
\addlinespace
Ministral-8B          & Logit (Geom.)   & 0.232 & 0.260 & 0.022 & 0.187 & 0.021 & 0.187 & 0.025 & 0.187 & 0.233 & 0.248 \\
                      & Logit (P-true)  & 0.349 & 0.313 & 0.032 & 0.187 & 0.031 & 0.187 & 0.038 & 0.188 & 0.245 & 0.251 \\
                      & Verbal          & 0.542 & 0.504 & 0.038 & \textbf{0.133} & 0.026 & \textbf{0.129} & 0.034 & \textbf{0.130} & 0.339 & 0.251 \\
\addlinespace
Qwen3.5-Flash         & Verbal          & 0.517 & 0.493 & 0.018 & 0.179 & 0.016 & 0.179 & 0.022 & 0.181 & 0.225 & 0.250 \\
\bottomrule
\end{tabular}%
}
\end{table}

\section{Soft Label Distillation for Calibrating BERT with LLM Annotations}

Directly prompting LLMs provides both a value prediction and a confidence score in one step. However, LLM generated confidence is often miscalibrated and varies significantly across different prompts. Furthermore, obtaining reliable human confidence annotations is costly. To address this challenge, we adopt a simple yet effective soft label distillation approach based on encoder models. Specifically, we use an LLM to produce explicit probability outputs, then distill these soft labels into a BERT model. This smoothing effect allows the student model to absorb uncertainty information from the teacher while producing more stable probability estimates. This mitigates the instability of raw LLM confidences, leading to lower ECE.

\paragraph{Module Design}
% Given an input sentence \(x\), the LLM produces a predicted score \(\hat{y}_{LLM}\) and verbal confidence \(c \in [0,100]\). We convert this pair into a \(K\)-class soft target (\(K=11\) for FOMC and \(K=10\) for other datasets): the class corresponding to \(\hat{y}_{LLM}\) receives probability \(c\), and the remaining classes share probability mass \(100-c\). We then fine-tune a BERT classifier on these soft targets and use its class probabilities as calibrated confidence proxies.

Given an input sentence \(x\), the LLM produces a predicted score \(\hat{y}_{LLM}\) and verbal confidence \(c \in [0,100]\). We convert this pair into a \(K\) class soft target (\(K=11\) for FOMC and \(K=10\) for other datasets). The class corresponding to \(\hat{y}_{LLM}\) receives probability \(c\), and the remaining classes share probability mass \(100-c\). We then finetune a BERT classifier on these soft targets and use its class probabilities as calibrated confidence proxies. This follows the soft label distillation intuition that teacher uncertainty can provide useful supervision beyond hard labels \citep{zhou2021rethinking}. Additional training details are provided in Appendix~\ref{sec:traing_details_dilstill}.

% \paragraph{Implementation Details}

% The data are randomly split into training and validation sets, at 80$\%$ and 20$\%$ respectively. We use a batch size with 16, a maximum sequence length of 512, a learning rate of \(2\times10^{-5}\), a weight decay of 0.01, linear warmup with 10$\%$ warmup steps, a KL divergence soft label loss with a temperature of 1.0, and gradient clipping at 1.0. Training runs for 30 epochs. For reported results, inference uses the epoch 30 checkpoint.

\paragraph{Experiments Results}

\begin{table}[htbp]
\centering
\caption{Paired calibration comparison per dataset. Lower is better for both ECE and Brier. The $\Delta$ columns report BERT minus LLM (GPT-5-nano Verbal), so negative values indicate improvement after soft label distillation.}
\label{tab:llm_bert_calibration_3dp}
\begin{tabular}{l ccc ccc}
\toprule
 & \multicolumn{3}{c}{\textbf{ECE} $\downarrow$} & \multicolumn{3}{c}{\textbf{Brier} $\downarrow$} \\
\cmidrule(lr){2-4} \cmidrule(lr){5-7}
\textbf{Dataset} & LLM & BERT & $\Delta$ & LLM & BERT & $\Delta$ \\
\midrule
Argument Quality   & 0.530 & 0.374 & $-0.156$ & 0.356 & 0.208 & $-0.148$ \\
EmoBank Arousal    & 0.488 & 0.330 & $-0.158$ & 0.378 & 0.234 & $-0.144$ \\
EmoBank Dominance  & 0.446 & 0.222 & $-0.225$ & 0.349 & 0.172 & $-0.176$ \\
EmoBank Valence    & 0.236 & 0.117 & $-0.119$ & 0.327 & 0.235 & $-0.092$ \\
Fomc               & 0.248 & 0.136 & $-0.113$ & 0.351 & 0.265 & $-0.087$ \\
Formality          & 0.384 & 0.175 & $-0.208$ & 0.377 & 0.260 & $-0.117$ \\
Hatespeech         & 0.659 & 0.468 & $-0.191$ & 0.539 & 0.363 & $-0.176$ \\
Humicroedit        & 0.270 & 0.128 & $-0.141$ & 0.320 & 0.248 & $-0.072$ \\
Politeness         & 0.365 & 0.106 & $-0.259$ & 0.391 & 0.249 & $-0.142$ \\
\midrule
Average            & 0.403 & 0.228 & $-0.174$ & 0.376 & 0.248 & $-0.128$ \\
\bottomrule
\end{tabular}
\end{table}

\begin{figure}
    \centering
    \includegraphics[width=1\linewidth]{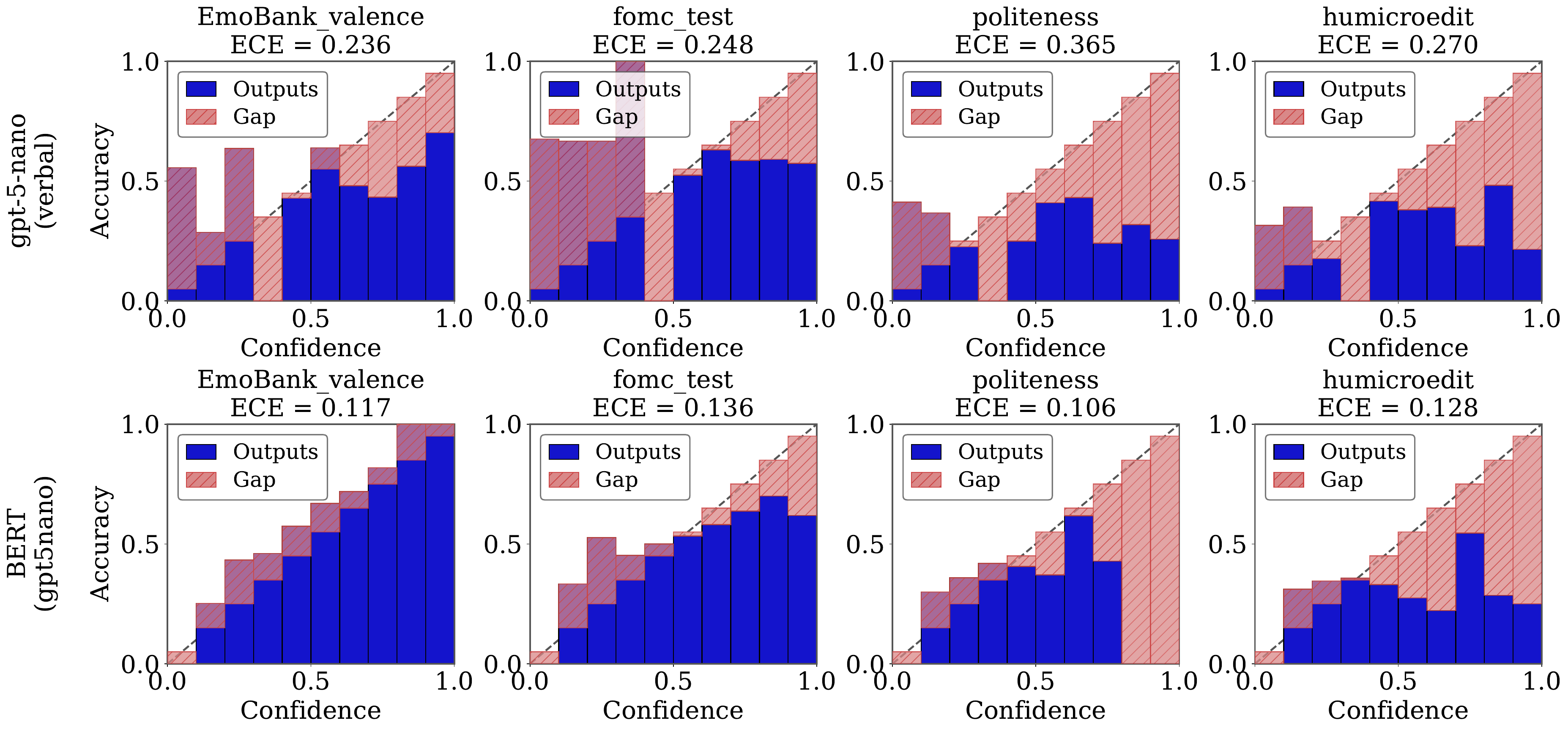}
    \caption{Bert method vs original GPT-5-nano Reliability Diagram.}
    \label{fig:bert_vs_gpt5nano01}
\end{figure}

Table~\ref{tab:llm_bert_calibration_3dp} shows gains from this distillation strategy. Averaged across datasets, ECE drops from 0.408 to 0.228, a relative reduction of about 43.4\%, and Brier drops from 0.376 to 0.248, a relative reduction of about 34\%. The FOMC is an informative boundary case. While ECE improves slightly, Brier worsens. Despite this challenging scenario, the cross dataset average and per-dataset deltas reinforce the reliability advantages of weak supervision distillation. Confidence soft labels regularise brittle one hot supervision, averaging student training over noisy teacher signals and suppressing prompt level variance. Probabilistic student outputs provide a smoother uncertainty surface for filtering. This interpretation is consistent with prior findings on weak supervision denoising and uncertainty aware soft label training\citep{ratner2017snorkel, Mukherjee2020UncertaintyawareSF,zhou2021rethinking}.

% Under the paired view, all datasets move in the same direction, indicating that the gain is systematic rather than a single dataset artifact. In practical terms, this means downstream confidence thresholding decisions are based on probability estimates that are materially closer to observed correctness.

\section{Related Work}

\subsection{LLMs as Measurement Tools in Social Science}

LLMs are increasingly deployed as scalable measurement instruments in social science, replacing or assisting manual coding for constructs such as stance, sentiment, and toxicity \citep{hansen2023fedspeak,bail2024can,asirvatham2026gpt}. This trend is expanding from labeling assistance to full empirical workflows \citep{asirvatham2026gpt,bail2024can}. Existing studies primarily report agreement with human labels or average predictive quality~\citep{asirvatham2026gpt,bail2024can}, which is necessary but insufficient when researchers perform confidence based filtering before downstream regressions. Without calibrated uncertainty, such threshold-based filtering risks either discarding valid data or retaining noisy estimates, directly distorting the subsequent statistical inferences ~\citep{angelopoulos2023prediction}. The missing piece is a systematic calibration audit for social science measurement, whether reported confidence reliably tracks measurement correctness under practical tolerance criteria. Our work fills this gap with a multi task empirical study and shows that calibration errors can propagate into materially different regression estimates. We further provide a simple confidence aware BERT distillation baseline to improve reliability in this setting.

% \subsection{Calibration Foundations}

\subsection{Calibration Methods for LLMs}

Calibration aligns predicted confidence with realized correctness, among predictions assigned confidence $p$, roughly a fraction $p$ should be correct \citep{kapoor2024calibrationtuning}. This reliability perspective is standard in uncertainty estimation, and ECE remains a widely used summary statistic \citep{guo2017calibration}. In NLP, prior work shows that strong task accuracy does not guarantee trustworthy confidence \citep{desai2020calibrating}. Recent LLM studies reinforce this point, overconfidence can persist after instruction tuning and alignment \citep{kapoor2024calibrationtuning,zhang2024ufcalibration}. These findings motivate our focus on confidence quality rather than accuracy alone \citep{huang2024rankcalibration}.

Recent work extends calibration analysis and intervention to LLMs, mostly in factual QA and classification tasks with binary correctness targets \citep{kadavath2022language,kapoor2024calibrationtuning,manggala2025qacalibration}. Representative methods include self reported confidence prompting \citep{kadavath2022language}, response agreement and prompt-style analyses \citep{xia2025influences}, fidelity aware decomposition of confidence \citep{zhang2024ufcalibration}, and activation level calibration for factual decoding \citep{liu2024actcab}. Rank based evaluation frameworks further argue that heterogeneous uncertainty scores should be compared through ordering consistency rather than a single shared numeric scale \citep{huang2024rankcalibration}. Now studies push this line to long context and multi answer regimes, showing that calibration failures persist and that single scalar confidence can be insufficient for claim level reliability \citep{muller2026benchmarking,wang2026mace,liu2026cure,jenane2026entropy}. Across these lines, a common conclusion is that raw LLM confidence is often misaligned with correctness and needs explicit correction \citep{kapoor2024calibrationtuning,zhang2024ufcalibration}. However, most evaluations remain QA and classification tasks \citep{manggala2025qacalibration,xia2025influences}, which transfer imperfectly to social science measurement tasks with subjective or tolerance based labels.

\section{Discussion}

Our findings expose a critical alignment calibration gap in LLM-based social science measurement, where high predictive alignment does not guarantee reliable absolute confidence. Furthermore, our evaluation reveals a pervasive resolution collapse among standard post-hoc calibration methods. While parametric scaling mathematically minimizes $T\text{-ECE}_\epsilon$, it frequently collapses predictions toward the dataset's base rate, which destroys discriminative resolution and yields uninformative estimates. To mitigate this, our soft label distillation approach successfully transfers relative uncertainty from LLMs into smaller encoders to provide a scalable and calibrated pipeline. 

Positive and Negative Impacts. By improving the reliability of these measurement pipelines, this work helps reduce confidence driven selection bias in computational social science. However, potential negative impacts remain if our framework is misused to justify excessive trust in automated scoring within sensitive domains such as political communication or toxicity moderation. Without sufficient human oversight, such applications risk amplifying unfair or domain specific biases.

Limitations and Future Work. Our evaluation is primarily restricted to English constructs and textual input. Future work should explore multilingual and multimodal calibration and investigate intrinsic alignment techniques to preserve confidence distribution integrity during pre-training.

\section{Contributions}

% To the best of our knowledge, this is the first comprehensive calibration audit of LLMs for continuous social science measurement. Our core contributions are \textbf{(1) Benchmarking the Calibration Gap:} We audit 14 datasets, demonstrating that state-of-the-art LLMs systematically misestimate measurement correctness despite high human alignment. \textbf{(2) Exposing ECE Collapse:} We prove that post-hoc calibration methods often game $T\text{-ECE}_\epsilon$ at the cost of resolution loss, establishing the necessity of dual-lens (ECE and Brier) evaluation. \textbf{(3) Distillation Baseline:} We introduce a soft label distillation framework that systematically improves both ECE and Brier scores, offering a reliable deployment pipeline.

This paper studies calibration as a core validity requirement for LLM-based social science measurement. Across 14 constructs, we show that strong commercial LLMs do not guarantee reliable confidence and that confidence filtering can distort downstream regression estimation in policy analyses. Our main findings are threefold. First, we provide a systematic dataset showing a persistent calibration gap across model families and social science tasks. Second, we show that optimizing a single calibration metric can hide resolution collapse, motivating evaluation with both $T\text{-ECE}_\epsilon$ and Brier score. Third, we present a lightweight soft-label distillation pipeline that improves calibration quality consistently across datasets, with average reductions in both ECE and Brier. Overall, our results suggest a practical guideline for social science workflows, report calibration metrics together with task performance, and avoid treating raw LLM confidence as directly trustworthy for threshold-based selection in downstream empirical estimation.

\medskip
\small
\bibliographystyle{plainnat}
\bibliography{main}

@article{asirvatham2026gpt,
  title={GPT as a Measurement Tool},
  author={Hemanth Asirvatham and Elliott Mokski and Andrei Vasiliev Dmitry Shleifer},
  journal={SSRN Electronic Journal},
  year={2026},
  url={https://api.semanticscholar.org/CorpusID:285809516}
}

@article{bail2024can,
  title={Can Generative AI improve social science?},
  author={Christopher A Bail},
  journal={Proceedings of the National Academy of Sciences of the United States of America},
  year={2024},
  volume={121},
  url={https://api.semanticscholar.org/CorpusID:269646697}
}

@article{hansen2023fedspeak,
  title={Can ChatGPT Decipher Fedspeak?},
  author={Anne Lundgaard Hansen and Sophia Kazinnik},
  journal={SSRN Electronic Journal},
  year={2023},
  url={https://api.semanticscholar.org/CorpusID:258039570}
}

@inproceedings{xia2025influences,
  title={Influences on LLM Calibration: A Study of Response Agreement, Loss Functions, and Prompt Styles},
  author={Yuxi Xia and Pedro Henrique Luz de Araujo and Klim Zaporojets and Benjamin Roth},
  booktitle={Annual Meeting of the Association for Computational Linguistics},
  year={2025},
  url={https://api.semanticscholar.org/CorpusID:275342783}
}

@inproceedings{kapoor2024calibrationtuning,
  title     = {Large Language Models Must Be Taught to Know What They Don't Know},
  author    = {Kapoor, Sanyam and Gruver, Nate and Roberts, Manley and Collins, Katherine and Pal, Arka and Bhatt, Umang and Weller, Adrian and Dooley, Samuel and Goldblum, Micah and Wilson, Andrew Gordon},
  booktitle = {Advances in Neural Information Processing Systems},
  year      = {2024}
}

@article{kadavath2022language,
  title={Language Models (Mostly) Know What They Know},
  author={Saurav Kadavath and Tom Conerly and Amanda Askell and Thomas Henighan and Dawn Drain and Ethan Perez and Nicholas Schiefer and Zachary Dodds and Nova Dassarma and Eli Tran-Johnson and Scott Johnston and Sheer El-Showk and Andy Jones and Nelson Elhage and Tristan Hume and Anna Chen and Yuntao Bai and Sam Bowman and Stanislav Fort and Deep Ganguli and Danny Hernandez and Josh Jacobson and John Kernion and Shauna Kravec and Liane Lovitt and Kamal Ndousse and Catherine Olsson and Sam Ringer and Dario Amodei and Tom B. Brown and Jack Clark and Nicholas Joseph and Benjamin Mann and Sam McCandlish and Chris Olah and Jared Kaplan},
  journal={ArXiv},
  year={2022},
  volume={abs/2207.05221},
  url={https://api.semanticscholar.org/CorpusID:250451161}
}

@inproceedings{guo2017calibration,
author = {Guo, Chuan and Pleiss, Geoff and Sun, Yu and Weinberger, Kilian Q.},
title = {On calibration of modern neural networks},
year = {2017},
publisher = {JMLR.org},
booktitle = {Proceedings of the 34th International Conference on Machine Learning - Volume 70},
pages = {1321–1330},
numpages = {10},
location = {Sydney, NSW, Australia},
series = {ICML'17}
}

@inproceedings{desai2020calibrating,
    title = "Calibration of Pre-trained Transformers",
    author = "Desai, Shrey  and
      Durrett, Greg",
    editor = "Webber, Bonnie  and
      Cohn, Trevor  and
      He, Yulan  and
      Liu, Yang",
    booktitle = "Proceedings of the 2020 Conference on Empirical Methods in Natural Language Processing (EMNLP)",
    month = nov,
    year = "2020",
    address = "Online",
    publisher = "Association for Computational Linguistics",
    url = "https://aclanthology.org/2020.emnlp-main.21/",
    doi = "10.18653/v1/2020.emnlp-main.21",
    pages = "295--302",
}

@inproceedings{zhang2024ufcalibration,
  title={Calibrating the Confidence of Large Language Models by Eliciting Fidelity},
  author={Mozhi Zhang and Mianqiu Huang and Rundong Shi and Linsen Guo and Chong Peng and Peng Yan and Yaqian Zhou and Xipeng Qiu},
  booktitle={Conference on Empirical Methods in Natural Language Processing},
  year={2024},
  url={https://api.semanticscholar.org/CorpusID:268876453}
}

@inproceedings{huang2024rankcalibration,
    title = "Uncertainty in Language Models: Assessment through Rank-Calibration",
    author = "Huang, Xinmeng  and
      Li, Shuo  and
      Yu, Mengxin  and
      Sesia, Matteo  and
      Hassani, Hamed  and
      Lee, Insup  and
      Bastani, Osbert  and
      Dobriban, Edgar",
    editor = "Al-Onaizan, Yaser  and
      Bansal, Mohit  and
      Chen, Yun-Nung",
    booktitle = "Proceedings of the 2024 Conference on Empirical Methods in Natural Language Processing",
    month = nov,
    year = "2024",
    address = "Miami, Florida, USA",
    publisher = "Association for Computational Linguistics",
    url = "https://aclanthology.org/2024.emnlp-main.18/",
    doi = "10.18653/v1/2024.emnlp-main.18",
    pages = "284--312",
}

@inproceedings{liu2024actcab,
  title={Enhancing Language Model Factuality via Activation-Based Confidence Calibration and Guided Decoding},
  author={Xin Liu and Farima Fatahi Bayat and Lu Wang},
  booktitle={Conference on Empirical Methods in Natural Language Processing},
  year={2024},
  url={https://api.semanticscholar.org/CorpusID:270620078}
}

@inproceedings{manggala2025qacalibration,
  title={QA-Calibration of Language Model Confidence Scores},
  author={Putra Manggala and Atalanti A. Mastakouri and Elke Kirschbaum and Shiva Prasad Kasiviswanathan and Aaditya Ramdas},
  booktitle={International Conference on Learning Representations},
  year={2024},
  url={https://api.semanticscholar.org/CorpusID:273228151}
}

@inproceedings{jenane2026entropy,
  title={From Entropy to Calibrated Uncertainty: Training Language Models to Reason About Uncertainty},
  author={Azza Jenane and Nassim Walha and Lukas Kuhn and Florian Buettner},
  year={2026},
  url={https://api.semanticscholar.org/CorpusID:286367709}
}

@inproceedings{liu2026cure,
  title={Think Through Uncertainty: Improving Long-Form Generation Factuality via Reasoning Calibration},
  author={Xin Liu and Lu Wang},
  year={2026},
  url={https://api.semanticscholar.org/CorpusID:287436193}
}

@article{muller2026benchmarking,
  title={Benchmarking Uncertainty Calibration in Large Language Model Long-Form Question Answering},
  author={Philip M{\"u}ller and Nicholas Popovic and Michael F{\"a}rber and P{\'e}ter Steinbach},
  journal={ArXiv},
  year={2026},
  volume={abs/2602.00279},
  url={https://api.semanticscholar.org/CorpusID:285270867}
}

@article{wang2026mace,
  title={Evaluating and Calibrating LLM Confidence on Questions with Multiple Correct Answers},
  author={Yuhang Wang and Shiyu Ni and Zhikai Ding and Zihang Zhan and Yuanzi Li and Keping Bi},
  journal={ArXiv},
  year={2026},
  volume={abs/2602.07842},
  url={https://api.semanticscholar.org/CorpusID:285452681}
}

@inproceedings{Shah2023TrillionDW,
  title={Trillion Dollar Words: A New Financial Dataset, Task \& Market Analysis},
  author={Agam Shah and Suvan Paturi and Sudheer Chava},
  booktitle={Annual Meeting of the Association for Computational Linguistics},
  year={2023},
  url={https://api.semanticscholar.org/CorpusID:258685646}
}

@article{xiong2023can,
  title={Can llms express their uncertainty? an empirical evaluation of confidence elicitation in llms},
  author={Xiong, Miao and Hu, Zhiyuan and Lu, Xinyang and Li, Yifei and Fu, Jie and He, Junxian and Hooi, Bryan},
  journal={arXiv preprint arXiv:2306.13063},
  year={2023}
}

@article{neuhaus1999bias,
  title={Bias and efficiency loss due to misclassified responses in binary regression},
  author={Neuhaus, John M},
  journal={Biometrika},
  volume={86},
  number={4},
  pages={843--855},
  year={1999},
  publisher={Oxford University Press}
}

@article{de2025study,
  title={A study of calibration as a measurement of trustworthiness of large language models in biomedical natural language processing},
  author={de Oliveira, Rodrigo and Garber, Matthew and Gwinnutt, James M and Rashidi, Emaan and Hwang, Jwu-Hsuan and Gilmour, William and Nanavati, Jay and Zine El Abidine, Khaldoun and Mack, Christina DeFilippo},
  journal={JAMIA open},
  volume={8},
  number={4},
  pages={ooaf058},
  year={2025},
  publisher={Oxford University Press}
}

@article{zhou2021rethinking,
  title={Rethinking Soft Labels for Knowledge Distillation: A Bias-Variance Tradeoff Perspective},
  author={Helong Zhou and Liangchen Song and Jiajie Chen and Ye Zhou and Guoli Wang and Junsong Yuan and Qian Zhang},
  journal={ArXiv},
  year={2021},
  volume={abs/2102.00650},
  url={https://api.semanticscholar.org/CorpusID:231740588}
}

@article{ratner2017snorkel,
  title={Snorkel: Rapid Training Data Creation with Weak Supervision},
  author={Alexander J. Ratner and Stephen H. Bach and Henry R. Ehrenberg and Jason Alan Fries and Sen Wu and Christopher R{\'e}},
  journal={Proceedings of the VLDB Endowment. International Conference on Very Large Data Bases},
  year={2017},
  volume={11 3},
  pages={
          269-282
        },
  url={https://api.semanticscholar.org/CorpusID:6730236}
}

@inproceedings{Mukherjee2020UncertaintyawareSF,
  title={Uncertainty-aware Self-training for Few-shot Text Classification},
  author={Subhabrata Mukherjee and Ahmed Hassan Awadallah},
  booktitle={Neural Information Processing Systems},
  year={2020},
  url={https://api.semanticscholar.org/CorpusID:227276483}
}

@inproceedings{tian2023just,
  title={Just ask for calibration: Strategies for eliciting calibrated confidence scores from language models fine-tuned with human feedback},
  author={Tian, Katherine and Mitchell, Eric and Zhou, Allan and Sharma, Archit and Rafailov, Rafael and Yao, Huaxiu and Finn, Chelsea and Manning, Christopher D},
  booktitle={Proceedings of the 2023 Conference on Empirical Methods in Natural Language Processing},
  pages={5433--5442},
  year={2023}
}

@article{liu2023litcab,
  title={Litcab: Lightweight language model calibration over short-and long-form responses},
  author={Liu, Xin and Khalifa, Muhammad and Wang, Lu},
  journal={arXiv preprint arXiv:2310.19208},
  year={2023}
}

@article{rathje2024gpt,
  title={GPT is an effective tool for multilingual psychological text analysis},
  author={Rathje, Steve and Mirea, Dan-Mircea and Sucholutsky, Ilia and Marjieh, Raja and Robertson, Claire E and Van Bavel, Jay J},
  journal={Proceedings of the National Academy of Sciences},
  volume={121},
  number={34},
  pages={e2308950121},
  year={2024},
  publisher={National Academy of Sciences}
}

@article{PavlickAndTetreault-2016:TACL,
  author =  {Ellie Pavlick and Joel Tetreault},
  title =   {An Empirical Analysis of Formality in Online Communication},
  journal = {Transactions of the Association for Computational Linguistics},
  year =    {2016},
  publisher = {Association for Computational Linguistics}
}

@inproceedings{danescu2013computational,
  title={A computational approach to politeness with application to social factors},
  author={Danescu-Niculescu-Mizil, Cristian and Sudhof, Moritz and Jurafsky, Dan and Leskovec, Jure and Potts, Christopher},
  booktitle={Proceedings of the 51st Annual Meeting of the Association for Computational Linguistics (Volume 1: Long Papers)},
  pages={250--259},
  year={2013}
}

@article{kennedy2020constructing,
  title={Constructing interval variables via faceted Rasch measurement and multitask deep learning: a hate speech application},
  author={Kennedy, Chris J and Bacon, Geoff and Sahn, Alexander and von Vacano, Claudia},
  journal={arXiv preprint arXiv:2009.10277},
  year={2020}
}

@article{hossain2019president,
  title={" President Vows to Cut< Taxes> Hair": Dataset and Analysis of Creative Text Editing for Humorous Headlines},
  author={Hossain, Nabil and Krumm, John and Gamon, Michael},
  journal={arXiv preprint arXiv:1906.00274},
  year={2019}
}

@inproceedings{buechel2017emobank,
  title={Emobank: Studying the impact of annotation perspective and representation format on dimensional emotion analysis},
  author={Buechel, Sven and Hahn, Udo},
  booktitle={Proceedings of the 15th Conference of the European Chapter of the Association for Computational Linguistics: Volume 2, Short Papers},
  pages={578--585},
  year={2017}
}

@inproceedings{ashida2022towards,
  title={Towards automatic generation of messages countering online hate speech and microaggressions},
  author={Ashida, Mana and Komachi, Mamoru},
  booktitle={Proceedings of the Sixth Workshop on Online Abuse and Harms (WOAH)},
  pages={11--23},
  year={2022}
}

@article{DBLP:journals/corr/abs-1911-11408,
  author       = {Shai Gretz and
                  Roni Friedman and
                  Edo Cohen{-}Karlik and
                  Assaf Toledo and
                  Dan Lahav and
                  Ranit Aharonov and
                  Noam Slonim},
  title        = {A Large-scale Dataset for Argument Quality Ranking: Construction and
                  Analysis},
  journal      = {CoRR},
  volume       = {abs/1911.11408},
  year         = {2019},
  url          = {http://arxiv.org/abs/1911.11408},
  eprinttype    = {arXiv},
  eprint       = {1911.11408},
  bibsource    = {dblp computer science bibliography, https://dblp.org}
}

@article{le2023uncovering,
  title={Uncovering the semantics of concepts using GPT-4},
  author={Le Mens, Ga{\"e}l and Kov{\'a}cs, Bal{\'a}zs and Hannan, Michael T and Pros, Guillem},
  journal={Proceedings of the National Academy of Sciences},
  volume={120},
  number={49},
  pages={e2309350120},
  year={2023},
  publisher={National Academy of Sciences}
}

@article{platt1999probabilistic,
  title={Probabilistic outputs for support vector machines and comparisons to regularized likelihood methods},
  author={Platt, John and others},
  journal={Advances in large margin classifiers},
  volume={10},
  number={3},
  pages={61--74},
  year={1999},
  publisher={Cambridge, MA}
}

@inproceedings{kull2017beta,
  title={Beta calibration: a well-founded and easily implemented improvement on logistic calibration for binary classifiers},
  author={Kull, Meelis and Silva Filho, Telmo and Flach, Peter},
  booktitle={Artificial intelligence and statistics},
  pages={623--631},
  year={2017},
  organization={PMLR}
}

@article{barlow1972isotonic,
  title={The isotonic regression problem and its dual},
  author={Barlow, Richard E and Brunk, Hugh D},
  journal={Journal of the American Statistical Association},
  volume={67},
  number={337},
  pages={140--147},
  year={1972},
  publisher={Taylor \& Francis}
}

@inproceedings{ding2021local,
  title={Local temperature scaling for probability calibration},
  author={Ding, Zhipeng and Han, Xu and Liu, Peirong and Niethammer, Marc},
  booktitle={Proceedings of the IEEE/CVF International Conference on Computer Vision},
  pages={6889--6899},
  year={2021}
}

@article{kumar2019verified,
  title={Verified uncertainty calibration},
  author={Kumar, Ananya and Liang, Percy S and Ma, Tengyu},
  journal={Advances in neural information processing systems},
  volume={32},
  year={2019}
}

@article{angelopoulos2023prediction,
  title={Prediction-powered inference},
  author={Angelopoulos, Anastasios N and Bates, Stephen and Fannjiang, Clara and Jordan, Michael I and Zrnic, Tijana},
  journal={Science},
  volume={382},
  number={6671},
  pages={669--674},
  year={2023},
  publisher={American Association for the Advancement of Science}
}

%%%%%%%%%%%%%%%%%%%%%%%%%%%%%%%%%%%%%%%%%%%%%%%%%%%%%%%%%%%%

\appendix

\section{Construct definitions} \label{sec:constructs_definition}

% \appendix

% \section{Detailed Construct Definitions}
% \label{sec:appendix_data}

All datasets used in our study are publicly available and designated for academic research. They are distributed under open licenses (e.g., MIT, CC BY 3.0/4.0, and CC BY-NC 4.0) or specific non-commercial research agreements (e.g., IBM Project Debater). Our usage complies with their respective terms and data redistribution policies. Table \ref{tab:appendix_constructs_part1} and Table \ref{tab:appendix_constructs_part2} provides the detailed prompts and construct definitions used for human annotation across all evaluated datasets. 

\begin{table}[htbp]
\centering
\caption{Datasets for Social Science Measurement. We report target constructs and our sampled size ($N$).}
\label{tab:dataset_summary}
\begin{tabular}{l l r}
\toprule
\textbf{Dataset} & \textbf{Construct(s)} & \textbf{Sampled $N$} \\
\midrule
CHASM & Offensiveness & 291 \\
FOMC Test Dataset & Dovish / Hawkish tone & 422 \\
Concept Alignment & Democrat, Republican, Mystery, Romance & 807  \\
Pavlick Formality  & Formality & 1{,}000 \\
Stanford Politeness & Politeness & 1{,}000 \\
Measuring Hate Speech & Hate speech & 1{,}000 \\
EmoBank & Valence, Arousal, Dominance & 1{,}000 \\
Argument Quality (AQ-30k) & Argument quality & 1{,}000 \\
Humicroedit & Funniness & 1{,}000 \\
\bottomrule
\end{tabular}
\end{table}

\begin{table}[htbp]
\centering
\caption{Detailed Construct Definitions and Annotation Prompts (Part I)}
\label{tab:appendix_constructs_part1}
\footnotesize
\begin{tabular}{p{2.8cm} p{2.2cm} p{8.5cm}}
\toprule
\textbf{Dataset} & \textbf{Construct} & \textbf{Prompt / Definition} \\
\midrule
\textbf{Pavlick Formality} & Formality & Rate the linguistic formality of the text. Near 0 means extremely informal, and near 100 means extremely formal, with intermediate values as appropriate. Use your intuitive judgment of formality and use the full ratings scale. \\
\midrule
\textbf{Stanford Politeness} & Politeness & Rate the politeness of the text. Near 0 means extremely impolite, and near 100 means extremely polite, with intermediate values as appropriate. Use your intuitive judgment of politeness and use the full ratings scale. \\
\midrule
\textbf{Measuring Hate Speech} & Hate speech & Rate the hate speech intensity of the text on a scale from 0 to 100. We linearly rescale the original continuous hate\_speech\_score to 0-100, where higher = more hateful and lower = less hateful. \\
\midrule
\textbf{CHASM} & Offensiveness & Rate how offensive the text is. Near 0 means not offensive, and near 100 means extremely offensive, with intermediate values as appropriate. Use your intuitive judgment and the full ratings scale. \\
\midrule
\textbf{EmoBank} & Valence & Rate the valence (pleasure) of the emotion evoked by the text. 0 means very negative/unhappy/annoyed, 100 means very positive/happy/satisfied, with intermediate values representing gradual transitions. Use your intuitive judgment and the full ratings scale. \\
\cmidrule{2-3}
 & Arousal & Rate the arousal (activation) of the emotion evoked by the text. 0 means very calm/relaxed/sleepy, 100 means very excited/nervous/aroused, with intermediate values representing gradual transitions. Use your intuitive judgment and the full ratings scale. \\
\cmidrule{2-3}
 & Dominance & Rate the dominance (control) of the emotion evoked by the text. 0 means feeling submissive/influenced/guided (low control), 100 means feeling dominant/in control/influential (high control), with intermediate values representing gradual transitions. Use your intuitive judgment and the full ratings scale. \\
\bottomrule
\end{tabular}
\end{table}

\begin{table}[htbp]
\centering
\caption{Detailed Construct Definitions and Annotation Prompts (Part II)}
\label{tab:appendix_constructs_part2}
\footnotesize
\begin{tabular}{p{2.8cm} p{2.2cm} p{8.5cm}}
\toprule
\textbf{Dataset} & \textbf{Construct} & \textbf{Prompt / Definition} \\
\midrule
\textbf{Argument Quality Ranking 30k} & Argument quality & Rate the overall quality of the argument on a scale from 0 to 100. 0 means very low quality (poor grammar, unclear, irrelevant, or unpersuasive), 100 means very high quality (clear, relevant, well-structured, and effective in supporting the stance), with intermediate values representing gradual transitions. Consider``Would you recommend using this argument as-is in a speech on the topic?'' Use your intuitive judgment and the full ratings scale. \\
\midrule
\textbf{Humicroedit} & Humor (funniness) & Grade the edited headline as funny if it would be funny to a large audience, regardless of your own stance toward the issue, entities, or information expressed in the headline. 0 means not funny at all, 100 means extremely funny, with intermediate values representing gradual transitions. Use your intuitive judgment and the full ratings scale. \\
\midrule
\textbf{FOMC Dataset} & Stance (Dovish / Hawkish) & Rate the stance of the text on a scale from 0 to 100. Near 0 means extremely dovish (easing monetary policy), near 100 means extremely hawkish (tightening monetary policy), and 50 represents a neutral stance. Use your intuitive judgment and the full ratings scale. \\
\midrule
\textbf{OSF Concept} & Mystery & Here's a DESCRIPTION ``TEXT.'' How typical is this DESCRIPTION of the Mystery concept? Provide your response as a score between 0 and 100 where 0 means ``Not typical at all'' and 100 means ``Extremely typical.'' \\
\cmidrule{2-3}
 & Romance & Here's a DESCRIPTION ``TEXT.'' How typical is this DESCRIPTION of the Romance concept? Provide your response as a score between 0 and 100 where 0 means ``Not typical at all'' and 100 means ``Extremely typical.'' \\
\cmidrule{2-3}
 & Democrat & Here's a DESCRIPTION ``TEXT.'' How typical is this DESCRIPTION of the Democratic Party concept? Provide your response as a score between 0 and 100 where 0 means ``Not typical at all'' and 100 means ``Extremely typical.'' \\
\cmidrule{2-3}
 & Republican & Here's a DESCRIPTION ``TEXT.'' How typical is this DESCRIPTION of the Republican Party concept? Provide your response as a score between 0 and 100 where 0 means ``Not typical at all'' and 100 means ``Extremely typical.'' \\
\bottomrule
\end{tabular}
\end{table}

\newpage

\section{Prompts}
\label{sec:prompt_appendix}

\subsection{Prompt Templates}
\label{sec:prompt_template}

Below is the base prompt template used for all social science measurement tasks. Dynamic fields (e.g., text, attributes, and definitions) are populated according to the specific dataset being evaluated.

\begin{quote}
\small
\texttt{[Input Content]} \\
\texttt{Your task: for each attribute below, rate how strongly the provided content manifests it.} \\

\texttt{BEGIN ATTRIBUTES} \\
\texttt{\{Attribute Names and Definitions\}} \\
\texttt{END ATTRIBUTES} \\

\texttt{BEGIN RATING SCALE} \\
\texttt{Use integers 0-100 (inclusive). low = absent, high = extreme, mid = moderate.} \\
\texttt{Use the full range and every increment, do not round to 5s/10s.} \\
\texttt{Extremes are rare, use near 0 only if truly absent and near 100 only if overwhelming.} \\
\texttt{Use moderate intermediates (e.g., 19, 67, 32) to account for nuance where applicable.} \\
\texttt{Aim for the rating that serves as the optimal center for a $\pm\epsilon$ point tolerance interval, ensuring the highest probability of capturing the true intensity.} \\
\texttt{END RATING SCALE} \\

\texttt{Method (per attribute) pick one exact integer. Stick to provided scale. Double check before choosing extremes. Interpret gradations as  absent$\rightarrow$faint$\rightarrow$moderate$\rightarrow$abundant$\rightarrow$extreme. Don't overlook subtlety, don't default to extremes. Consider full spectrum, including intermediate gradations. High accuracy/precision is critical, it needs deep, holistic analysis of content.} \\

\texttt{Rules:} \\
\texttt{- Judge each attribute independently and separately from each other} \\
\texttt{- Absolutely no indirect inference from other attributes or cross attribute leakage} \\
\texttt{- Only measure the direct signal of each attribute alone in the content, NOT what is implied by other attributes, CRUCIAL each attribute measured independently on its own direct, specifically relevant signal} \\

\texttt{Output JSON only, in following format:} \\
\texttt{\{} \\
\texttt{\quad "attribute\_name" rating,} \\
\texttt{\quad ...} \\
\texttt{\}}
\label{tab:prompt_template_base}
\end{quote}

\subsection{Verbal Construct}
\label{sec:verbal_construct}

\begin{quote}
    \texttt{"accuracy\_estimate"f"What is the probability (0-100) that your prediction is within $\pm$\{tolerance\_threshold\} points of the true value?"}
\end{quote}

\section{Implementation and Inference Settings}
\label{sec:inference_settings}

Local model inference was conducted on NVIDIA L20 and GeForce RTX 3090 GPUs using the Hugging Face library in bfloat16 precision. For API-based models, we utilized the OpenAI and OpenRouter platforms. To ensure the full capturing of probabilistic distributions for calibration analysis, all inference was performed with a temperature of 1.0, a top\_p of 1.0, and a maximum completion limit of 4,096 tokens.

\section{Training Details for Soft Label Distillation}

\label{sec:traing_details_dilstill}
The data are randomly split into training and validation sets, at 80$\%$ and 20$\%$ respectively. We use a batch size of 16, a maximum sequence length of 512, a learning rate of \(2\times10^{-5}\), a weight decay of 0.01, linear warmup with 10$\%$ warmup steps, a KL divergence soft label loss with a temperature of 1.0, and gradient clipping at 1.0. Training runs for 30 epochs. For reported results, inference uses the epoch 30 checkpoint. We trained the model on a single NVIDIA L20 GPU, and the process took about 30 minutes for each dataset.

\section{Full Results}
\subsection{Observing Miscalibration}
\label{sec: observing_miscalibration}

\begin{table}[h]
\centering
\caption{Calibration performance on EmoBank-arousal and Humicroedit datasets. Each method is evaluated on ECE, Brier score, and MH metrics (lower is better).}
\label{tab:emobank_arousal_humicroedit}
\resizebox{\textwidth}{!}{%
\small
\begin{tabular}{@{}ll*{3}{ccc}@{}}
\toprule
\textbf{Model} & \textbf{Method} & \multicolumn{3}{c}{\textbf{EmoBank-arousal}} & \multicolumn{3}{c}{\textbf{Humicroedit}} \\
\cmidrule(lr){3-5} \cmidrule(lr){6-8}
 & & \textbf{ECE} $\downarrow$ & \textbf{Brier} $\downarrow$ & \textbf{MH} $\uparrow$ & \textbf{ECE} $\downarrow$ & \textbf{Brier} $\downarrow$ & \textbf{MH} $\uparrow$ \\
\midrule
DeepSeek-R1-7B        & Logit (Geom.)   & 0.334 & 0.219 & 0.185 & 0.194 & 0.232 & 0.126 \\
                      & Logit (P-true)  & 0.819 & 0.767 & 0.185 & 0.655 & 0.640 & 0.126 \\
                      & Verbal          & 0.464 & 0.396 & 0.284 & 0.316 & 0.337 & 0.050 \\
\addlinespace
Qwen2.5-7B            & Logit (Geom.)   & 0.194 & 0.269 & 0.321 & 0.228 & 0.261 & 0.145 \\
                      & Logit (P-true)  & 0.626 & 0.598 & 0.321 & 0.640 & 0.617 & 0.145 \\
                      & Verbal          & 0.475 & 0.434 & 0.295 & 0.413 & 0.405 & 0.134 \\
\addlinespace
DeepSeek-V3.2         & Verbal          & 0.534 & 0.455 & 0.391 & 0.507 & 0.482 & 0.195 \\
\addlinespace
Gemma-2-9B            & Logit (Geom.)   & 0.331 & 0.244 & 0.298 & 0.279 & 0.262 & 0.160 \\
                      & Logit (P-true)  & 0.548 & 0.428 & 0.298 & 0.647 & 0.625 & 0.160 \\
                      & Verbal          & 0.557 & 0.491 & 0.310 & 0.458 & 0.426 & 0.179 \\
\addlinespace
GPT-5-mini            & Verbal          & 0.668 & 0.576 & 0.455 & 0.329 & 0.337 & 0.370 \\
GPT-5-nano            & Verbal          & 0.488 & 0.378 & 0.387 & 0.270 & 0.320 & 0.289 \\
\addlinespace
Ministral-8B          & Logit (Geom.)   & 0.270 & 0.259 & 0.183 & 0.217 & 0.266 & 0.035 \\
                      & Logit (P-true)  & 0.391 & 0.320 & 0.183 & 0.262 & 0.288 & 0.035 \\
                      & Verbal          & 0.627 & 0.589 & 0.386 & 0.623 & 0.576 & 0.179 \\
\addlinespace
Qwen3.5-Flash         & Verbal          & 0.642 & 0.574 & 0.469 & 0.369 & 0.387 & 0.313 \\
\bottomrule
\end{tabular}%
}
\end{table}

\begin{table}[htbp]
\centering
\caption{Calibration performance on OSF-concept-republican and EmoBank-dominance datasets.}
\label{tab:osf_republican_emobank_dominance}
\resizebox{\textwidth}{!}{%
\small
\begin{tabular}{@{}ll*{3}{ccc}@{}}
\toprule
\textbf{Model} & \textbf{Method} & \multicolumn{3}{c}{\textbf{OSF-concept-republican}} & \multicolumn{3}{c}{\textbf{EmoBank-dominance}} \\
\cmidrule(lr){3-5} \cmidrule(lr){6-8}
 & & \textbf{ECE} $\downarrow$ & \textbf{Brier} $\downarrow$ & \textbf{MH} $\uparrow$ & \textbf{ECE} $\downarrow$ & \textbf{Brier} $\downarrow$ & \textbf{MH} $\uparrow$ \\
\midrule
DeepSeek-R1-7B        & Logit (Geom.)   & 0.254 & 0.213 & 0.158 & 0.361 & 0.210 & 0.032 \\
                      & Logit (P-true)  & 0.729 & 0.696 & 0.158 & 0.838 & 0.774 & 0.032 \\
                      & Verbal          & 0.553 & 0.520 & 0.098 & 0.418 & 0.357 & 0.055 \\
\addlinespace
Qwen2.5-7B            & Logit (Geom.)   & 0.157 & 0.260 & 0.466 & 0.149 & 0.260 & 0.182 \\
                      & Logit (P-true)  & 0.542 & 0.540 & 0.466 & 0.572 & 0.547 & 0.182 \\
                      & Verbal          & 0.526 & 0.502 & 0.529 & 0.402 & 0.404 & 0.092 \\
\addlinespace
DeepSeek-V3.2         & Verbal          & 0.397 & 0.405 & 0.733 & 0.466 & 0.420 & 0.234 \\
\addlinespace
Gemma-2-9B            & Logit (Geom.)   & 0.143 & 0.260 & 0.499 & 0.268 & 0.245 & 0.071 \\
                      & Logit (P-true)  & 0.409 & 0.416 & 0.499 & 0.537 & 0.456 & 0.071 \\
                      & Verbal          & 0.427 & 0.420 & 0.515 & 0.530 & 0.486 & 0.100 \\
\addlinespace
GPT-5-mini            & Verbal          & 0.418 & 0.408 & 0.775 & 0.624 & 0.527 & 0.273 \\
GPT-5-nano            & Verbal          & 0.278 & 0.321 & 0.645 & 0.446 & 0.348 & 0.093 \\
\addlinespace
Ministral-8B          & Logit (Geom.)   & 0.230 & 0.261 & 0.296 & 0.190 & 0.247 & 0.043 \\
                      & Logit (P-true)  & 0.392 & 0.338 & 0.296 & 0.327 & 0.307 & 0.043 \\
                      & Verbal          & 0.550 & 0.518 & 0.701 & 0.647 & 0.584 & 0.160 \\
\addlinespace
Qwen3.5-Flash         & Verbal          & 0.502 & 0.481 & 0.748 & 0.523 & 0.481 & 0.250 \\
\bottomrule
\end{tabular}%
}
\end{table}

\begin{table}[htbp]
\centering
\caption{Calibration performance on OSF-concept-democrat and FOMC-test datasets.}
\label{tab:osf_democrat_fomc}
\resizebox{\textwidth}{!}{%
\small
\begin{tabular}{@{}ll*{3}{ccc}@{}}
\toprule
\textbf{Model} & \textbf{Method} & \multicolumn{3}{c}{\textbf{OSF-concept-democrat}} & \multicolumn{3}{c}{\textbf{FOMC-test}} \\
\cmidrule(lr){3-5} \cmidrule(lr){6-8}
 & & \textbf{ECE} $\downarrow$ & \textbf{Brier} $\downarrow$ & \textbf{MH} $\uparrow$ & \textbf{ECE} $\downarrow$ & \textbf{Brier} $\downarrow$ & \textbf{MH} $\uparrow$ \\
\midrule
DeepSeek-R1-7B        & Logit (Geom.)   & 0.228 & 0.221 & 0.191 & 0.030 & 0.247 & 0.010 \\
                      & Logit (P-true)  & 0.699 & 0.666 & 0.191 & 0.483 & 0.474 & 0.010 \\
                      & Verbal          & 0.489 & 0.474 & 0.229 & 0.339 & 0.395 & 0.076 \\
\addlinespace
Qwen2.5-7B            & Logit (Geom.)   & 0.129 & 0.252 & 0.558 & 0.063 & 0.245 & 0.172 \\
                      & Logit (P-true)  & 0.526 & 0.520 & 0.558 & 0.532 & 0.528 & 0.172 \\
                      & Verbal          & 0.513 & 0.491 & 0.602 & 0.355 & 0.380 & 0.117 \\
\addlinespace
DeepSeek-V3.2         & Verbal          & 0.294 & 0.336 & 0.744 & 0.170 & 0.282 & 0.391 \\
\addlinespace
Gemma-2-9B            & Logit (Geom.)   & 0.158 & 0.259 & 0.330 & 0.049 & 0.244 & 0.204 \\
                      & Logit (P-true)  & 0.404 & 0.392 & 0.330 & 0.352 & 0.355 & 0.204 \\
                      & Verbal          & 0.366 & 0.368 & 0.349 & 0.348 & 0.387 & 0.271 \\
\addlinespace
GPT-5-mini            & Verbal          & 0.376 & 0.393 & 0.794 & 0.222 & 0.276 & 0.480 \\
GPT-5-nano            & Verbal          & 0.271 & 0.326 & 0.548 & 0.248 & 0.351 & 0.384 \\
\addlinespace
Ministral-8B          & Logit (Geom.)   & 0.189 & 0.263 & 0.256 & 0.140 & 0.257 & 0.123 \\
                      & Logit (P-true)  & 0.332 & 0.316 & 0.256 & 0.290 & 0.311 & 0.123 \\
                      & Verbal          & 0.510 & 0.482 & 0.712 & 0.444 & 0.437 & 0.133 \\
\addlinespace
Qwen3.5-Flash         & Verbal          & 0.458 & 0.457 & 0.746 & 0.371 & 0.397 & 0.520 \\
\bottomrule
\end{tabular}%
}
\end{table}

\begin{table}[htbp]
\centering
\caption{Calibration performance on OSF-concept-mystery and Argument-quality datasets.}
\label{tab:osf_mystery_argument}
\resizebox{\textwidth}{!}{%
\small
\begin{tabular}{@{}ll*{3}{ccc}@{}}
\toprule
\textbf{Model} & \textbf{Method} & \multicolumn{3}{c}{\textbf{OSF-concept-mystery}} & \multicolumn{3}{c}{\textbf{Argument-quality}} \\
\cmidrule(lr){3-5} \cmidrule(lr){6-8}
 & & \textbf{ECE} $\downarrow$ & \textbf{Brier} $\downarrow$ & \textbf{MH} $\uparrow$ & \textbf{ECE} $\downarrow$ & \textbf{Brier} $\downarrow$ & \textbf{MH} $\uparrow$ \\
\midrule
DeepSeek-R1-7B        & Logit (Geom.)   & 0.151 & 0.227 & 0.420 & 0.190 & 0.238 & 0.165 \\
                      & Logit (P-true)  & 0.697 & 0.691 & 0.420 & 0.665 & 0.637 & 0.165 \\
                      & Verbal          & 0.589 & 0.556 & 0.488 & 0.308 & 0.308 & 0.168 \\
\addlinespace
Qwen2.5-7B            & Logit (Geom.)   & 0.118 & 0.264 & 0.749 & 0.397 & 0.245 & 0.113 \\
                      & Logit (P-true)  & 0.501 & 0.507 & 0.749 & 0.799 & 0.753 & 0.113 \\
                      & Verbal          & 0.538 & 0.514 & 0.575 & 0.543 & 0.439 & 0.209 \\
\addlinespace
DeepSeek-V3.2         & Verbal          & 0.351 & 0.364 & 0.786 & 0.686 & 0.552 & 0.269 \\
\addlinespace
Gemma-2-9B            & Logit (Geom.)   & 0.161 & 0.258 & 0.746 & 0.408 & 0.260 & 0.244 \\
                      & Logit (P-true)  & 0.498 & 0.494 & 0.746 & 0.662 & 0.580 & 0.244 \\
                      & Verbal          & 0.503 & 0.474 & 0.661 & 0.667 & 0.553 & 0.252 \\
\addlinespace
GPT-5-mini            & Verbal          & 0.468 & 0.446 & 0.868 & 0.736 & 0.582 & 0.320 \\
GPT-5-nano            & Verbal          & 0.221 & 0.305 & 0.815 & 0.530 & 0.355 & 0.268 \\
\addlinespace
Ministral-8B          & Logit (Geom.)   & 0.254 & 0.273 & 0.343 & 0.343 & 0.253 & 0.115 \\
                      & Logit (P-true)  & 0.429 & 0.380 & 0.343 & 0.422 & 0.294 & 0.115 \\
                      & Verbal          & 0.618 & 0.588 & 0.706 & 0.546 & 0.449 & 0.273 \\
\addlinespace
Qwen3.5-Flash         & Verbal          & 0.569 & 0.542 & 0.873 & 0.765 & 0.657 & 0.187 \\
\bottomrule
\end{tabular}%
}
\end{table}

\begin{table}[htbp]
\centering
\caption{Calibration performance on CHASM-offensive and Hatespeech datasets.}
\label{tab:chasm_hatespeech}
\resizebox{\textwidth}{!}{%
\small
\begin{tabular}{@{}ll*{3}{ccc}@{}}
\toprule
\textbf{Model} & \textbf{Method} & \multicolumn{3}{c}{\textbf{CHASM-offensive}} & \multicolumn{3}{c}{\textbf{Hatespeech}} \\
\cmidrule(lr){3-5} \cmidrule(lr){6-8}
 & & \textbf{ECE} $\downarrow$ & \textbf{Brier} $\downarrow$ & \textbf{MH} $\uparrow$ & \textbf{ECE} $\downarrow$ & \textbf{Brier} $\downarrow$ & \textbf{MH} $\uparrow$ \\
\midrule
DeepSeek-R1-7B        & Logit (Geom.)   & 0.255 & 0.226 & 0.337 & 0.369 & 0.226 & 0.528 \\
                      & Logit (P-true)  & 0.745 & 0.717 & 0.337 & 0.820 & 0.769 & 0.528 \\
                      & Verbal          & 0.297 & 0.332 & 0.340 & 0.481 & 0.414 & 0.606 \\
\addlinespace
Qwen2.5-7B            & Logit (Geom.)   & 0.249 & 0.270 & 0.484 & 0.336 & 0.288 & 0.628 \\
                      & Logit (P-true)  & 0.702 & 0.681 & 0.484 & 0.750 & 0.726 & 0.628 \\
                      & Verbal          & 0.354 & 0.391 & 0.520 & 0.549 & 0.497 & 0.660 \\
\addlinespace
DeepSeek-V3.2         & Verbal          & 0.277 & 0.345 & 0.538 & 0.574 & 0.501 & 0.750 \\
\addlinespace
Gemma-2-9B            & Logit (Geom.)   & 0.146 & 0.263 & 0.581 & 0.423 & 0.281 & 0.689 \\
                      & Logit (P-true)  & 0.287 & 0.284 & 0.581 & 0.735 & 0.683 & 0.689 \\
                      & Verbal          & 0.463 & 0.488 & 0.548 & 0.588 & 0.496 & 0.693 \\
\addlinespace
GPT-5-mini            & Verbal          & 0.077 & 0.167 & 0.670 & 0.780 & 0.700 & 0.627 \\
GPT-5-nano            & Verbal          & 0.181 & 0.290 & 0.603 & 0.659 & 0.539 & 0.607 \\
\addlinespace
Ministral-8B          & Logit (Geom.)   & 0.251 & 0.257 & 0.436 & 0.407 & 0.303 & 0.506 \\
                      & Logit (P-true)  & 0.380 & 0.316 & 0.436 & 0.452 & 0.325 & 0.506 \\
                      & Verbal          & 0.444 & 0.448 & 0.534 & 0.519 & 0.472 & 0.667 \\
\addlinespace
Qwen3.5-Flash         & Verbal          & 0.532 & 0.524 & 0.617 & 0.698 & 0.644 & 0.748 \\
\bottomrule
\end{tabular}%
}
\end{table}

\begin{table}[htbp]
\centering
\caption{Calibration performance on EmoBank-valence and Formality datasets.}
\label{tab:emobank_valence_formality}
\resizebox{\textwidth}{!}{%
\small
\begin{tabular}{@{}ll*{3}{ccc}@{}}
\toprule
\textbf{Model} & \textbf{Method} & \multicolumn{3}{c}{\textbf{EmoBank-valence}} & \multicolumn{3}{c}{\textbf{Formality}} \\
\cmidrule(lr){3-5} \cmidrule(lr){6-8}
 & & \textbf{ECE} $\downarrow$ & \textbf{Brier} $\downarrow$ & \textbf{MH} $\uparrow$ & \textbf{ECE} $\downarrow$ & \textbf{Brier} $\downarrow$ & \textbf{MH} $\uparrow$ \\
\midrule
DeepSeek-R1-7B        & Logit (Geom.)   & 0.145 & 0.233 & 0.363 & 0.138 & 0.240 & 0.333 \\
                      & Logit (P-true)  & 0.647 & 0.621 & 0.363 & 0.639 & 0.611 & 0.333 \\
                      & Verbal          & 0.428 & 0.438 & 0.426 & 0.373 & 0.365 & 0.344 \\
\addlinespace
Qwen2.5-7B            & Logit (Geom.)   & 0.076 & 0.260 & 0.573 & 0.183 & 0.255 & 0.445 \\
                      & Logit (P-true)  & 0.505 & 0.491 & 0.573 & 0.646 & 0.614 & 0.445 \\
                      & Verbal          & 0.394 & 0.404 & 0.570 & 0.457 & 0.441 & 0.468 \\
\addlinespace
DeepSeek-V3.2         & Verbal          & 0.290 & 0.351 & 0.710 & 0.380 & 0.395 & 0.603 \\
\addlinespace
Gemma-2-9B            & Logit (Geom.)   & 0.105 & 0.256 & 0.505 & 0.122 & 0.253 & 0.549 \\
                      & Logit (P-true)  & 0.459 & 0.438 & 0.505 & 0.509 & 0.501 & 0.549 \\
                      & Verbal          & 0.400 & 0.404 & 0.451 & 0.431 & 0.428 & 0.503 \\
\addlinespace
GPT-5-mini            & Verbal          & 0.316 & 0.353 & 0.796 & 0.431 & 0.418 & 0.746 \\
GPT-5-nano            & Verbal          & 0.236 & 0.327 & 0.720 & 0.384 & 0.377 & 0.479 \\
\addlinespace
Ministral-8B          & Logit (Geom.)   & 0.122 & 0.250 & 0.342 & 0.156 & 0.244 & 0.243 \\
                      & Logit (P-true)  & 0.268 & 0.296 & 0.342 & 0.313 & 0.314 & 0.243 \\
                      & Verbal          & 0.500 & 0.485 & 0.584 & 0.457 & 0.461 & 0.611 \\
\addlinespace
Qwen3.5-Flash         & Verbal          & 0.340 & 0.378 & 0.799 & 0.451 & 0.448 & 0.708 \\
\bottomrule
\end{tabular}%
}
\end{table}

\begin{table}[htbp]
\centering
\caption{Calibration performance on OSF-concept-romance and Politeness datasets.}
\label{tab:osf_romance_politeness}
\resizebox{\textwidth}{!}{%
\small
\begin{tabular}{@{}ll*{3}{ccc}@{}}
\toprule
\textbf{Model} & \textbf{Method} & \multicolumn{3}{c}{\textbf{OSF-concept-romance}} & \multicolumn{3}{c}{\textbf{Politeness}} \\
\cmidrule(lr){3-5} \cmidrule(lr){6-8}
 & & \textbf{ECE} $\downarrow$ & \textbf{Brier} $\downarrow$ & \textbf{MH} $\uparrow$ & \textbf{ECE} $\downarrow$ & \textbf{Brier} $\downarrow$ & \textbf{MH} $\uparrow$ \\
\midrule
DeepSeek-R1-7B        & Logit (Geom.)   & 0.045 & 0.237 & 0.636 & 0.209 & 0.238 & 0.186 \\
                      & Logit (P-true)  & 0.574 & 0.575 & 0.636 & 0.689 & 0.670 & 0.186 \\
                      & Verbal          & 0.512 & 0.507 & 0.683 & 0.377 & 0.374 & 0.164 \\
\addlinespace
Qwen2.5-7B            & Logit (Geom.)   & 0.116 & 0.274 & 0.753 & 0.119 & 0.267 & 0.385 \\
                      & Logit (P-true)  & 0.486 & 0.477 & 0.753 & 0.529 & 0.518 & 0.385 \\
                      & Verbal          & 0.532 & 0.501 & 0.754 & 0.382 & 0.380 & 0.326 \\
\addlinespace
DeepSeek-V3.2         & Verbal          & 0.346 & 0.373 & 0.833 & 0.272 & 0.333 & 0.337 \\
\addlinespace
Gemma-2-9B            & Logit (Geom.)   & 0.104 & 0.262 & 0.787 & 0.266 & 0.263 & 0.287 \\
                      & Logit (P-true)  & 0.496 & 0.492 & 0.787 & 0.637 & 0.604 & 0.287 \\
                      & Verbal          & 0.410 & 0.418 & 0.786 & 0.543 & 0.498 & 0.310 \\
\addlinespace
GPT-5-mini            & Verbal          & 0.388 & 0.386 & 0.879 & 0.690 & 0.592 & 0.423 \\
GPT-5-nano            & Verbal          & 0.211 & 0.290 & 0.841 & 0.365 & 0.391 & 0.334 \\
\addlinespace
Ministral-8B          & Logit (Geom.)   & 0.183 & 0.261 & 0.482 & 0.244 & 0.257 & 0.153 \\
                      & Logit (P-true)  & 0.325 & 0.331 & 0.482 & 0.414 & 0.356 & 0.153 \\
                      & Verbal          & 0.555 & 0.543 & 0.822 & 0.516 & 0.483 & 0.328 \\
\addlinespace
Qwen3.5-Flash         & Verbal          & 0.514 & 0.494 & 0.858 & 0.487 & 0.467 & 0.383 \\
\bottomrule
\end{tabular}%
}
\end{table}

\newpage

\subsection{Post-hoc Calibration Reliability Diagram}
\label{sec:post_hoc_calibration_reliability_diagram}

\begin{figure}[htbp]
    \centering
    \includegraphics[width=1\linewidth]{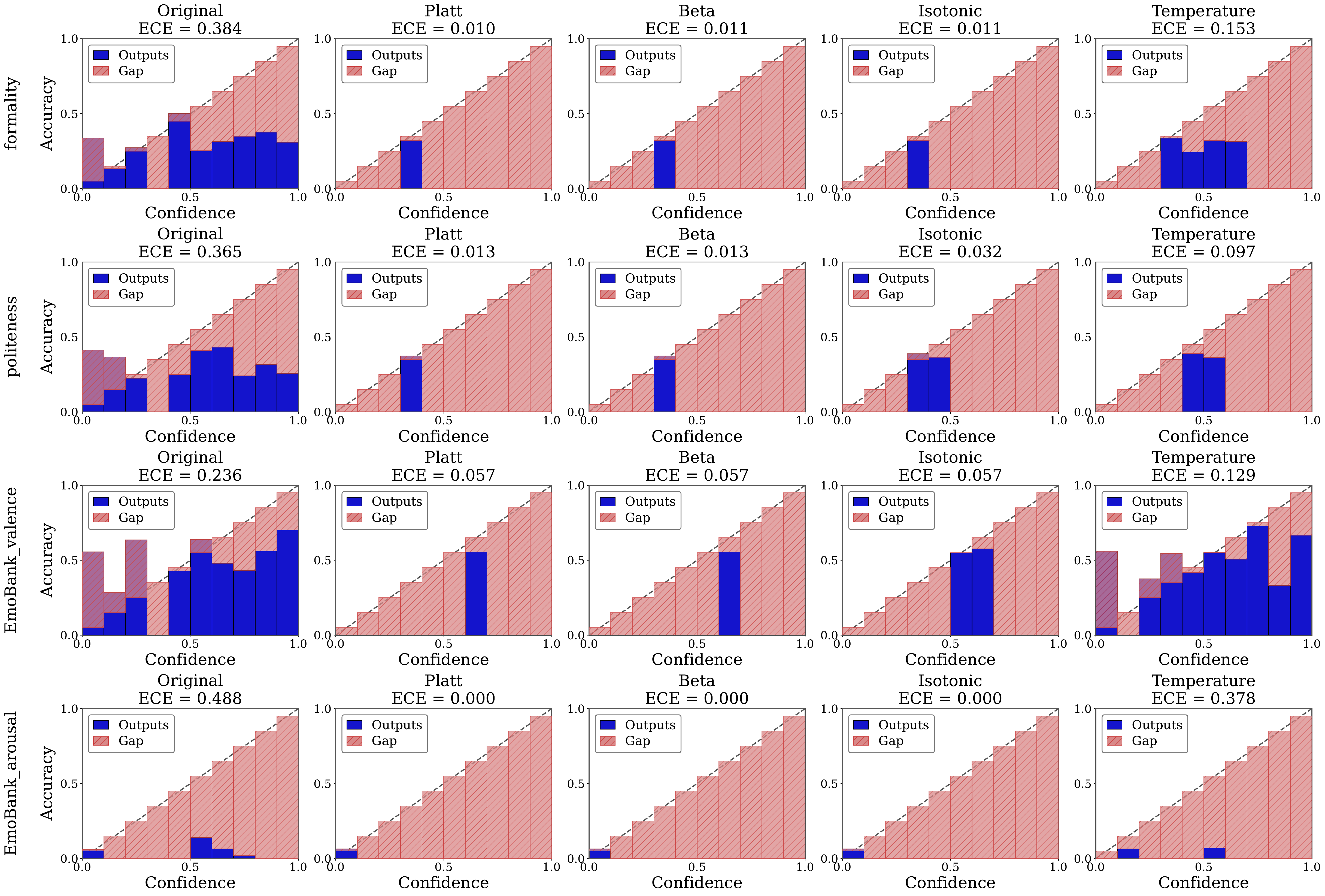}
    \caption{Post-hoc Reliability Diagram of datasets: formality, politeness, EmoBank\_valence, Emobank\_arousal.}
    \label{fig:posthoc_reliability_gpt5nano_part1}
\end{figure}

\begin{figure}[htbp]
    \centering
    \includegraphics[width=1\linewidth]{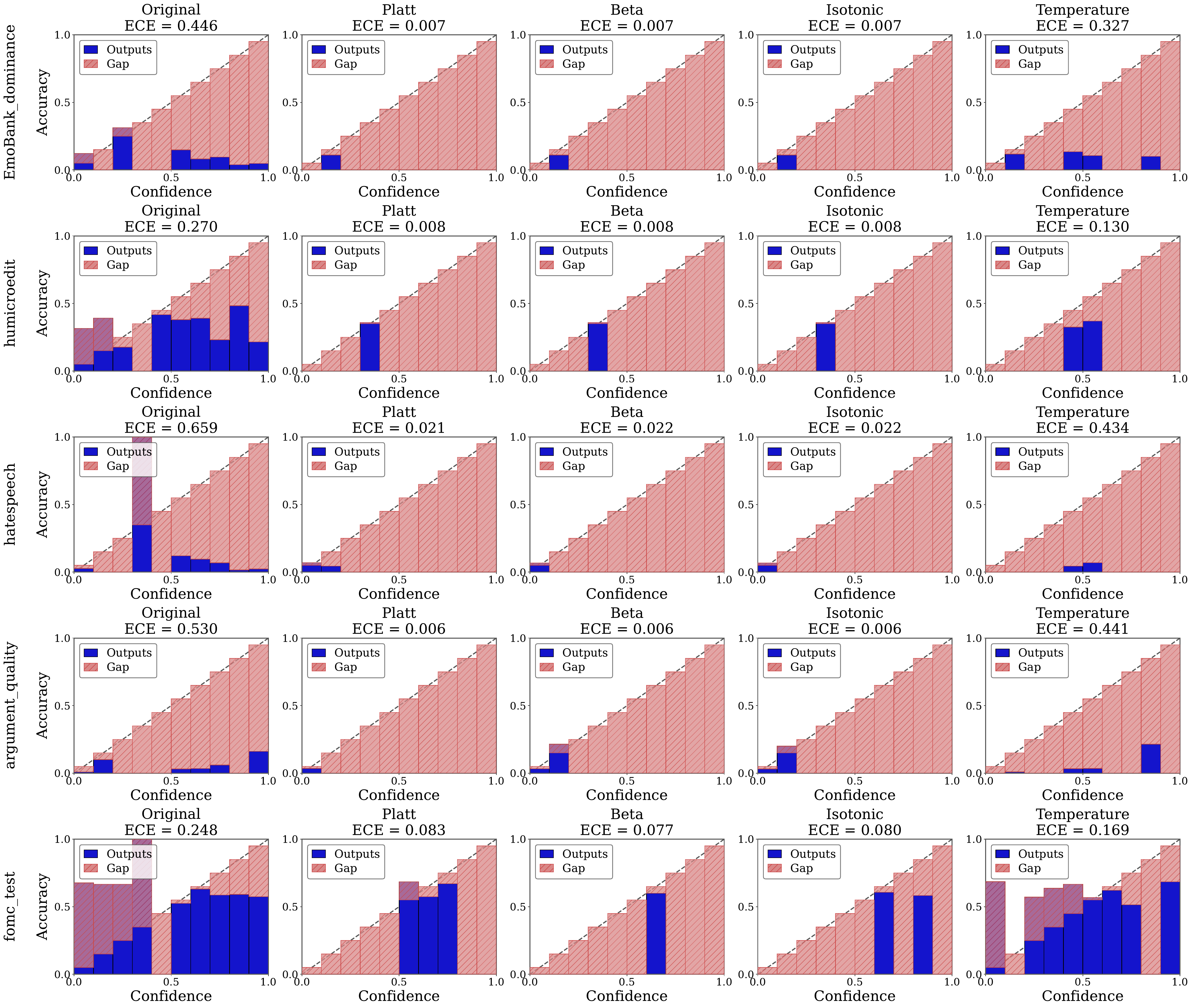}
    \caption{Post-hoc Reliability Diagram of datasets: EmoBank\_dominance, Humicredit, Hatespeech, Argument\_quality, Fomc\_test.}
    \label{fig:posthoc_reliability_gpt5nano_part2}
\end{figure}

\newpage

\subsection{Post-hoc Full Results}
\label{sec:post_hoc_calibration_results}

\begin{table}[htbp]
\centering
\caption{Calibration performance on EmoBank-arousal dataset. Each calibration method is trained on a dedicated 1,000 sample stratified subset.}
\label{tab:emobank_arousal}
\resizebox{\textwidth}{!}{%
\small
\begin{tabular}{@{}ll*{5}{cc}@{}}
\toprule
\textbf{Model} & \textbf{Method} & \multicolumn{2}{c}{\textbf{Original}} & \multicolumn{2}{c}{\textbf{Platt}} & \multicolumn{2}{c}{\textbf{Beta}} & \multicolumn{2}{c}{\textbf{Isotonic}} & \multicolumn{2}{c}{\textbf{Temperature}} \\
\cmidrule(lr){3-4} \cmidrule(lr){5-6} \cmidrule(lr){7-8} \cmidrule(lr){9-10} \cmidrule(l){11-12}
 & & \textbf{ECE} $\downarrow$ & \textbf{Brier} $\downarrow$ & \textbf{ECE} $\downarrow$ & \textbf{Brier} $\downarrow$ & \textbf{ECE} $\downarrow$ & \textbf{Brier} $\downarrow$ & \textbf{ECE} $\downarrow$ & \textbf{Brier} $\downarrow$ & \textbf{ECE} $\downarrow$ & \textbf{Brier} $\downarrow$ \\
\midrule
DeepSeek-R1-7B        & Logit (Geom.)   & 0.334 & 0.219 & 0.019 & 0.103 & 0.001 & 0.104 & 0.001 & 0.105 & \textbf{0.211} & \textbf{0.191} \\
                      & Logit (P-true)  & 0.819 & 0.767 & 0.007 & 0.103 & 0.016 & 0.103 & 0.003 & 0.103 & 0.391 & 0.252 \\
                      & Verbal          & 0.464 & 0.396 & 0.020 & 0.089 & 0.019 & 0.090 & 0.019 & 0.090 & 0.398 & 0.248 \\
\addlinespace
Qwen2.5-7B            & Logit (Geom.)   & \textbf{0.194} & 0.269 & 0.050 & 0.230 & 0.047 & 0.230 & 0.064 & 0.233 & 0.170 & 0.250 \\
                      & Logit (P-true)  & 0.626 & 0.598 & 0.047 & 0.230 & 0.047 & 0.230 & 0.049 & 0.230 & 0.175 & 0.252 \\
                      & Verbal          & 0.475 & 0.434 & 0.025 & 0.227 & 0.025 & 0.230 & 0.025 & 0.229 & 0.176 & 0.251 \\
\addlinespace
DeepSeek-V3.2         & Verbal          & 0.534 & 0.455 & 0.028 & 0.161 & 0.028 & 0.161 & 0.033 & 0.162 & 0.313 & 0.251 \\
\addlinespace
Gemma-2-9B            & Logit (Geom.)   & 0.331 & 0.244 & 0.020 & 0.127 & 0.019 & 0.128 & 0.024 & 0.128 & 0.342 & 0.244 \\
                      & Logit (P-true)  & 0.548 & 0.428 & 0.026 & 0.123 & 0.031 & 0.124 & 0.017 & 0.122 & 0.362 & 0.251 \\
                      & Verbal          & 0.557 & 0.491 & 0.012 & 0.142 & 0.012 & 0.142 & 0.011 & 0.142 & 0.332 & 0.247 \\
\addlinespace
GPT-5-mini            & Verbal          & 0.668 & 0.576 & 0.035 & 0.107 & 0.017 & 0.119 & 0.017 & 0.119 & 0.372 & 0.251 \\
GPT-5-nano            & Verbal          & 0.488 & 0.378 & \textbf{0.000} & 0.063 & \textbf{0.000} & 0.063 & \textbf{0.000} & 0.062 & 0.378 & 0.225 \\
\addlinespace
Ministral-8B          & Logit (Geom.)   & 0.270 & 0.259 & 0.016 & 0.174 & 0.015 & 0.173 & 0.015 & 0.174 & 0.282 & 0.249 \\
                      & Logit (P-true)  & 0.391 & 0.320 & 0.031 & 0.171 & 0.032 & 0.171 & 0.031 & 0.171 & 0.285 & 0.251 \\
                      & Verbal          & 0.627 & 0.589 & 0.023 & 0.098 & 0.023 & 0.098 & 0.024 & 0.098 & 0.397 & 0.251 \\
\addlinespace
Qwen3.5-Flash         & Verbal          & 0.642 & 0.574 & 0.006 & 0.108 & 0.006 & 0.108 & 0.010 & 0.109 & 0.378 & 0.249 \\
\bottomrule
\end{tabular}%
}
\end{table}

\begin{table}[htbp]
\centering
\caption{Calibration performance on Humicroedit dataset.}
\label{tab:humicroedit}
\resizebox{\textwidth}{!}{%
\small
\begin{tabular}{@{}ll*{5}{cc}@{}}
\toprule
\textbf{Model} & \textbf{Method} & \multicolumn{2}{c}{\textbf{Original}} & \multicolumn{2}{c}{\textbf{Platt}} & \multicolumn{2}{c}{\textbf{Beta}} & \multicolumn{2}{c}{\textbf{Isotonic}} & \multicolumn{2}{c}{\textbf{Temperature}} \\
\cmidrule(lr){3-4} \cmidrule(lr){5-6} \cmidrule(lr){7-8} \cmidrule(lr){9-10} \cmidrule(l){11-12}
 & & \textbf{ECE} $\downarrow$ & \textbf{Brier} $\downarrow$ & \textbf{ECE} $\downarrow$ & \textbf{Brier} $\downarrow$ & \textbf{ECE} $\downarrow$ & \textbf{Brier} $\downarrow$ & \textbf{ECE} $\downarrow$ & \textbf{Brier} $\downarrow$ & \textbf{ECE} $\downarrow$ & \textbf{Brier} $\downarrow$ \\
\midrule
DeepSeek-R1-7B        & Logit (Geom.)   & 0.194 & 0.232 & 0.004 & 0.191 & 0.003 & 0.190 & 0.003 & 0.190 & \textbf{0.131} & \textbf{0.224} \\
                      & Logit (P-true)  & 0.655 & 0.640 & 0.006 & 0.189 & 0.003 & 0.190 & 0.014 & 0.192 & 0.249 & 0.252 \\
                      & Verbal          & 0.316 & 0.337 & 0.019 & 0.190 & 0.019 & 0.190 & 0.019 & 0.190 & 0.244 & 0.248 \\
\addlinespace
Qwen2.5-7B            & Logit (Geom.)   & 0.228 & 0.261 & 0.015 & 0.202 & 0.015 & 0.202 & 0.018 & 0.203 & 0.223 & 0.250 \\
                      & Logit (P-true)  & 0.640 & 0.617 & 0.018 & 0.204 & 0.015 & 0.202 & 0.015 & 0.202 & 0.227 & 0.252 \\
                      & Verbal          & 0.413 & 0.405 & 0.006 & 0.227 & 0.006 & 0.226 & 0.007 & 0.226 & 0.163 & 0.251 \\
\addlinespace
DeepSeek-V3.2         & Verbal          & 0.507 & 0.482 & 0.023 & 0.198 & 0.010 & 0.198 & 0.014 & 0.199 & 0.231 & 0.251 \\
\addlinespace
Gemma-2-9B            & Logit (Geom.)   & 0.279 & 0.262 & 0.012 & 0.175 & 0.011 & 0.174 & 0.011 & 0.175 & 0.276 & 0.250 \\
                      & Logit (P-true)  & 0.647 & 0.625 & 0.043 & 0.172 & 0.011 & 0.174 & 0.021 & 0.178 & 0.279 & 0.252 \\
                      & Verbal          & 0.458 & 0.426 & 0.006 & 0.171 & 0.002 & 0.171 & 0.002 & 0.171 & 0.261 & 0.240 \\
\addlinespace
GPT-5-mini            & Verbal          & 0.329 & 0.337 & 0.031 & 0.235 & 0.030 & 0.235 & 0.028 & 0.235 & 0.136 & 0.251 \\
GPT-5-nano            & Verbal          & 0.270 & 0.320 & 0.008 & 0.233 & 0.008 & 0.233 & 0.008 & 0.233 & 0.130 & 0.248 \\
\addlinespace
Ministral-8B          & Logit (Geom.)   & 0.217 & 0.266 & 0.015 & 0.209 & 0.006 & 0.209 & 0.009 & 0.208 & 0.207 & 0.250 \\
                      & Logit (P-true)  & 0.262 & 0.288 & 0.008 & 0.209 & 0.006 & 0.209 & 0.019 & 0.210 & 0.208 & 0.250 \\
                      & Verbal          & 0.623 & 0.576 & 0.007 & 0.092 & 0.006 & 0.092 & 0.006 & 0.092 & 0.399 & 0.251 \\
\addlinespace
Qwen3.5-Flash         & Verbal          & 0.369 & 0.387 & 0.030 & 0.239 & 0.006 & 0.238 & 0.011 & 0.239 & 0.114 & 0.250 \\
\bottomrule
\end{tabular}%
}
\end{table}

\begin{table}[htbp]
\centering
\caption{Calibration performance on EmoBank-dominance dataset.}
\label{tab:emobank_dominance}
\resizebox{\textwidth}{!}{%
\small
\begin{tabular}{@{}ll*{5}{cc}@{}}
\toprule
\textbf{Model} & \textbf{Method} & \multicolumn{2}{c}{\textbf{Original}} & \multicolumn{2}{c}{\textbf{Platt}} & \multicolumn{2}{c}{\textbf{Beta}} & \multicolumn{2}{c}{\textbf{Isotonic}} & \multicolumn{2}{c}{\textbf{Temperature}} \\
\cmidrule(lr){3-4} \cmidrule(lr){5-6} \cmidrule(lr){7-8} \cmidrule(lr){9-10} \cmidrule(l){11-12}
 & & \textbf{ECE} $\downarrow$ & \textbf{Brier} $\downarrow$ & \textbf{ECE} $\downarrow$ & \textbf{Brier} $\downarrow$ & \textbf{ECE} $\downarrow$ & \textbf{Brier} $\downarrow$ & \textbf{ECE} $\downarrow$ & \textbf{Brier} $\downarrow$ & \textbf{ECE} $\downarrow$ & \textbf{Brier} $\downarrow$ \\
\midrule
DeepSeek-R1-7B        & Logit (Geom.)   & 0.361 & 0.210 & 0.012 & 0.071 & 0.012 & 0.072 & 0.012 & 0.072 & \textbf{0.224} & \textbf{0.166} \\
                      & Logit (P-true)  & 0.838 & 0.774 & 0.018 & 0.070 & 0.025 & 0.071 & 0.012 & 0.070 & 0.426 & 0.252 \\
                      & Verbal          & 0.418 & 0.357 & 0.002 & 0.075 & 0.002 & 0.075 & 0.002 & 0.075 & 0.403 & 0.240 \\
\addlinespace
Qwen2.5-7B            & Logit (Geom.)   & 0.149 & 0.260 & 0.005 & 0.237 & 0.006 & 0.237 & 0.007 & 0.236 & 0.143 & 0.250 \\
                      & Logit (P-true)  & 0.572 & 0.547 & 0.017 & 0.238 & 0.006 & 0.237 & 0.006 & 0.237 & 0.148 & 0.251 \\
                      & Verbal          & 0.402 & 0.404 & 0.016 & 0.245 & 0.016 & 0.245 & 0.048 & 0.247 & 0.106 & 0.251 \\
\addlinespace
DeepSeek-V3.2         & Verbal          & 0.466 & 0.420 & 0.022 & 0.202 & 0.021 & 0.202 & 0.021 & 0.201 & 0.239 & 0.251 \\
\addlinespace
Gemma-2-9B            & Logit (Geom.)   & 0.268 & 0.245 & 0.007 & 0.169 & 0.006 & 0.169 & 0.007 & 0.168 & 0.281 & 0.244 \\
                      & Logit (P-true)  & 0.537 & 0.456 & 0.029 & 0.166 & 0.034 & 0.166 & 0.014 & 0.165 & 0.301 & 0.251 \\
                      & Verbal          & 0.530 & 0.486 & 0.003 & 0.158 & 0.002 & 0.158 & 0.030 & 0.158 & 0.302 & 0.246 \\
\addlinespace
GPT-5-mini            & Verbal          & 0.624 & 0.527 & 0.005 & 0.130 & 0.002 & 0.132 & 0.011 & 0.133 & 0.351 & 0.251 \\
GPT-5-nano            & Verbal          & 0.446 & 0.348 & 0.007 & 0.102 & 0.007 & 0.102 & 0.007 & 0.102 & 0.327 & 0.226 \\
\addlinespace
Ministral-8B          & Logit (Geom.)   & 0.190 & 0.247 & 0.011 & 0.204 & 0.013 & 0.204 & 0.013 & 0.204 & 0.202 & 0.244 \\
                      & Logit (P-true)  & 0.327 & 0.307 & 0.027 & 0.204 & 0.022 & 0.204 & 0.022 & 0.205 & 0.220 & 0.251 \\
                      & Verbal          & 0.647 & 0.584 & 0.010 & 0.070 & 0.010 & 0.070 & 0.009 & 0.071 & 0.433 & 0.251 \\
\addlinespace
Qwen3.5-Flash         & Verbal          & 0.523 & 0.481 & 0.008 & 0.173 & 0.013 & 0.173 & 0.021 & 0.173 & 0.283 & 0.249 \\
\bottomrule
\end{tabular}%
}
\end{table}

\begin{table}[htbp]
\centering
\caption{Calibration performance on FOMC-test dataset.}
\label{tab:fomc_test}
\resizebox{\textwidth}{!}{%
\small
\begin{tabular}{@{}ll*{5}{cc}@{}}
\toprule
\textbf{Model} & \textbf{Method} & \multicolumn{2}{c}{\textbf{Original}} & \multicolumn{2}{c}{\textbf{Platt}} & \multicolumn{2}{c}{\textbf{Beta}} & \multicolumn{2}{c}{\textbf{Isotonic}} & \multicolumn{2}{c}{\textbf{Temperature}} \\
\cmidrule(lr){3-4} \cmidrule(lr){5-6} \cmidrule(lr){7-8} \cmidrule(lr){9-10} \cmidrule(l){11-12}
 & & \textbf{ECE} $\downarrow$ & \textbf{Brier} $\downarrow$ & \textbf{ECE} $\downarrow$ & \textbf{Brier} $\downarrow$ & \textbf{ECE} $\downarrow$ & \textbf{Brier} $\downarrow$ & \textbf{ECE} $\downarrow$ & \textbf{Brier} $\downarrow$ & \textbf{ECE} $\downarrow$ & \textbf{Brier} $\downarrow$ \\
\midrule
DeepSeek-R1-7B        & Logit (Geom.)   & 0.030 & 0.247 & 0.124 & 0.261 & 0.110 & 0.261 & 0.138 & 0.272 & \textbf{0.029} & \textbf{0.247} \\
                      & Logit (P-true)  & 0.483 & 0.474 & 0.104 & 0.264 & 0.101 & 0.258 & 0.101 & 0.258 & 0.085 & 0.254 \\
                      & Verbal          & 0.339 & 0.395 & 0.054 & 0.255 & 0.106 & 0.266 & 0.100 & 0.278 & 0.090 & 0.265 \\
\addlinespace
Qwen2.5-7B            & Logit (Geom.)   & 0.063 & 0.245 & 0.119 & 0.259 & 0.120 & 0.258 & 0.173 & 0.274 & 0.076 & 0.248 \\
                      & Logit (P-true)  & 0.532 & 0.528 & 0.179 & 0.282 & 0.156 & 0.270 & 0.201 & 0.288 & 0.167 & 0.267 \\
                      & Verbal          & 0.355 & 0.380 & 0.044 & 0.250 & 0.040 & 0.252 & 0.059 & 0.258 & 0.036 & 0.252 \\
\addlinespace
DeepSeek-V3.2         & Verbal          & 0.170 & 0.282 & 0.031 & 0.244 & 0.032 & 0.243 & 0.093 & 0.252 & 0.053 & 0.249 \\
\addlinespace
Gemma-2-9B            & Logit (Geom.)   & 0.049 & 0.244 & 0.040 & 0.243 & 0.034 & 0.245 & 0.075 & 0.252 & 0.049 & 0.244 \\
                      & Logit (P-true)  & 0.352 & 0.355 & 0.065 & 0.241 & 0.087 & 0.231 & 0.071 & 0.242 & 0.094 & 0.250 \\
                      & Verbal          & 0.348 & 0.387 & 0.053 & 0.260 & 0.079 & 0.276 & 0.104 & 0.287 & 0.031 & 0.251 \\
\addlinespace
GPT-5-mini            & Verbal          & 0.222 & 0.276 & 0.069 & 0.232 & 0.073 & 0.233 & 0.085 & 0.246 & 0.030 & 0.229 \\
GPT-5-nano            & Verbal          & 0.248 & 0.351 & 0.083 & 0.246 & 0.077 & 0.246 & 0.080 & 0.254 & 0.169 & 0.328 \\
\addlinespace
Ministral-8B          & Logit (Geom.)   & 0.140 & 0.257 & 0.027 & 0.229 & 0.022 & 0.228 & 0.037 & 0.230 & 0.130 & 0.249 \\
                      & Logit (P-true)  & 0.290 & 0.311 & 0.021 & 0.228 & 0.023 & 0.227 & 0.079 & 0.232 & 0.152 & 0.250 \\
                      & Verbal          & 0.444 & 0.437 & 0.204 & 0.287 & 0.100 & 0.249 & 0.162 & 0.264 & 0.112 & 0.252 \\
\addlinespace
Qwen3.5-Flash         & Verbal          & 0.371 & 0.397 & 0.086 & 0.256 & 0.082 & 0.253 & 0.122 & 0.268 & 0.051 & 0.252 \\
\bottomrule
\end{tabular}%
}
\end{table}

\begin{table}[htbp]
\centering
\caption{Calibration performance on Argument-quality dataset.}
\label{tab:argument_quality}
\resizebox{\textwidth}{!}{%
\small
\begin{tabular}{@{}ll*{5}{cc}@{}}
\toprule
\textbf{Model} & \textbf{Method} & \multicolumn{2}{c}{\textbf{Original}} & \multicolumn{2}{c}{\textbf{Platt}} & \multicolumn{2}{c}{\textbf{Beta}} & \multicolumn{2}{c}{\textbf{Isotonic}} & \multicolumn{2}{c}{\textbf{Temperature}} \\
\cmidrule(lr){3-4} \cmidrule(lr){5-6} \cmidrule(lr){7-8} \cmidrule(lr){9-10} \cmidrule(l){11-12}
 & & \textbf{ECE} $\downarrow$ & \textbf{Brier} $\downarrow$ & \textbf{ECE} $\downarrow$ & \textbf{Brier} $\downarrow$ & \textbf{ECE} $\downarrow$ & \textbf{Brier} $\downarrow$ & \textbf{ECE} $\downarrow$ & \textbf{Brier} $\downarrow$ & \textbf{ECE} $\downarrow$ & \textbf{Brier} $\downarrow$ \\
\midrule
DeepSeek-R1-7B        & Logit (Geom.)   & 0.190 & 0.238 & 0.018 & 0.199 & 0.015 & 0.199 & 0.032 & 0.201 & \textbf{0.155} & \textbf{0.237} \\
                      & Logit (P-true)  & 0.665 & 0.637 & 0.010 & 0.195 & 0.040 & 0.181 & 0.034 & 0.179 & 0.236 & 0.252 \\
                      & Verbal          & 0.308 & 0.308 & 0.041 & 0.197 & 0.031 & 0.198 & 0.037 & 0.199 & 0.230 & 0.250 \\
\addlinespace
Qwen2.5-7B            & Logit (Geom.)   & 0.397 & 0.245 & 0.001 & 0.088 & 0.003 & 0.088 & 0.004 & 0.087 & 0.403 & 0.247 \\
                      & Logit (P-true)  & 0.799 & 0.753 & 0.001 & 0.088 & 0.001 & 0.088 & 0.004 & 0.088 & 0.411 & 0.252 \\
                      & Verbal          & 0.543 & 0.439 & 0.017 & 0.116 & 0.028 & 0.116 & 0.022 & 0.117 & 0.372 & 0.251 \\
\addlinespace
DeepSeek-V3.2         & Verbal          & 0.686 & 0.552 & 0.004 & 0.068 & 0.004 & 0.068 & 0.005 & 0.068 & 0.429 & 0.251 \\
\addlinespace
Gemma-2-9B            & Logit (Geom.)   & 0.408 & 0.260 & \textbf{0.000} & 0.081 & \textbf{0.000} & 0.081 & 0.006 & 0.081 & 0.413 & 0.250 \\
                      & Logit (P-true)  & 0.662 & 0.580 & 0.001 & 0.081 & \textbf{0.000} & 0.081 & 0.018 & 0.082 & 0.417 & 0.252 \\
                      & Verbal          & 0.667 & 0.553 & 0.011 & 0.087 & 0.011 & 0.087 & 0.016 & 0.088 & 0.409 & 0.251 \\
\addlinespace
GPT-5-mini            & Verbal          & 0.736 & 0.582 & 0.009 & 0.039 & 0.009 & 0.039 & 0.010 & 0.039 & 0.462 & 0.251 \\
GPT-5-nano            & Verbal          & 0.530 & 0.355 & 0.006 & 0.034 & 0.006 & 0.033 & 0.006 & 0.033 & 0.441 & 0.236 \\
\addlinespace
Ministral-8B          & Logit (Geom.)   & 0.343 & 0.253 & 0.028 & 0.119 & 0.030 & 0.119 & 0.029 & 0.120 & 0.353 & 0.247 \\
                      & Logit (P-true)  & 0.422 & 0.294 & 0.031 & 0.118 & 0.030 & 0.118 & 0.029 & 0.117 & 0.366 & 0.251 \\
                      & Verbal          & 0.546 & 0.449 & 0.015 & 0.104 & 0.014 & 0.104 & 0.016 & 0.105 & 0.385 & 0.251 \\
\addlinespace
Qwen3.5-Flash         & Verbal          & 0.765 & 0.657 & \textbf{0.000} & 0.052 & \textbf{0.000} & 0.052 & 0.007 & 0.053 & 0.448 & 0.251 \\
\bottomrule
\end{tabular}%
}
\end{table}

\begin{table}[htbp]
\centering
\caption{Calibration performance on Hatespeech dataset.}
\label{tab:hatespeech}
\resizebox{\textwidth}{!}{%
\small
\begin{tabular}{@{}ll*{5}{cc}@{}}
\toprule
\textbf{Model} & \textbf{Method} & \multicolumn{2}{c}{\textbf{Original}} & \multicolumn{2}{c}{\textbf{Platt}} & \multicolumn{2}{c}{\textbf{Beta}} & \multicolumn{2}{c}{\textbf{Isotonic}} & \multicolumn{2}{c}{\textbf{Temperature}} \\
\cmidrule(lr){3-4} \cmidrule(lr){5-6} \cmidrule(lr){7-8} \cmidrule(lr){9-10} \cmidrule(l){11-12}
 & & \textbf{ECE} $\downarrow$ & \textbf{Brier} $\downarrow$ & \textbf{ECE} $\downarrow$ & \textbf{Brier} $\downarrow$ & \textbf{ECE} $\downarrow$ & \textbf{Brier} $\downarrow$ & \textbf{ECE} $\downarrow$ & \textbf{Brier} $\downarrow$ & \textbf{ECE} $\downarrow$ & \textbf{Brier} $\downarrow$ \\
\midrule
DeepSeek-R1-7B        & Logit (Geom.)   & 0.369 & 0.226 & 0.016 & 0.082 & 0.008 & 0.082 & 0.010 & 0.083 & \textbf{0.260} & \textbf{0.203} \\
                      & Logit (P-true)  & 0.820 & 0.769 & 0.008 & 0.082 & 0.008 & 0.082 & 0.009 & 0.083 & 0.412 & 0.252 \\
                      & Verbal          & 0.481 & 0.414 & 0.031 & 0.052 & 0.029 & 0.052 & 0.029 & 0.052 & 0.447 & 0.250 \\
\addlinespace
Qwen2.5-7B            & Logit (Geom.)   & 0.336 & 0.288 & 0.018 & 0.165 & 0.018 & 0.166 & 0.021 & 0.166 & 0.301 & 0.250 \\
                      & Logit (P-true)  & 0.750 & 0.726 & 0.018 & 0.166 & 0.027 & 0.167 & 0.032 & 0.167 & 0.304 & 0.253 \\
                      & Verbal          & 0.549 & 0.497 & 0.054 & 0.170 & 0.055 & 0.170 & 0.056 & 0.170 & 0.295 & 0.251 \\
\addlinespace
DeepSeek-V3.2         & Verbal          & 0.574 & 0.501 & 0.019 & 0.140 & 0.017 & 0.140 & 0.017 & 0.140 & 0.340 & 0.251 \\
\addlinespace
Gemma-2-9B            & Logit (Geom.)   & 0.423 & 0.281 & 0.007 & 0.091 & 0.001 & 0.091 & 0.015 & 0.091 & 0.402 & 0.250 \\
                      & Logit (P-true)  & 0.735 & 0.683 & 0.056 & 0.087 & 0.001 & 0.091 & 0.001 & 0.091 & 0.403 & 0.252 \\
                      & Verbal          & 0.588 & 0.496 & 0.010 & 0.073 & 0.009 & 0.073 & 0.009 & 0.073 & 0.405 & 0.245 \\
\addlinespace
GPT-5-mini            & Verbal          & 0.780 & 0.700 & 0.032 & 0.074 & 0.002 & 0.084 & 0.002 & 0.084 & 0.415 & 0.252 \\
GPT-5-nano            & Verbal          & 0.659 & 0.539 & 0.021 & 0.064 & 0.022 & 0.064 & 0.022 & 0.064 & 0.434 & 0.252 \\
\addlinespace
Ministral-8B          & Logit (Geom.)   & 0.407 & 0.303 & 0.026 & 0.115 & 0.026 & 0.115 & 0.027 & 0.116 & 0.369 & 0.250 \\
                      & Logit (P-true)  & 0.452 & 0.325 & 0.026 & 0.115 & 0.030 & 0.115 & 0.037 & 0.116 & 0.370 & 0.251 \\
                      & Verbal          & 0.519 & 0.472 & 0.014 & 0.038 & 0.013 & 0.038 & 0.014 & 0.038 & 0.462 & 0.250 \\
\addlinespace
Qwen3.5-Flash         & Verbal          & 0.698 & 0.644 & 0.006 & 0.088 & 0.003 & 0.088 & 0.003 & 0.088 & 0.404 & 0.251 \\
\bottomrule
\end{tabular}%
}
\end{table}

\begin{table}[htbp]
\centering
\caption{Calibration performance on EmoBank-valence dataset.}
\label{tab:emobank_valence}
\resizebox{\textwidth}{!}{%
\small
\begin{tabular}{@{}ll*{5}{cc}@{}}
\toprule
\textbf{Model} & \textbf{Method} & \multicolumn{2}{c}{\textbf{Original}} & \multicolumn{2}{c}{\textbf{Platt}} & \multicolumn{2}{c}{\textbf{Beta}} & \multicolumn{2}{c}{\textbf{Isotonic}} & \multicolumn{2}{c}{\textbf{Temperature}} \\
\cmidrule(lr){3-4} \cmidrule(lr){5-6} \cmidrule(lr){7-8} \cmidrule(lr){9-10} \cmidrule(l){11-12}
 & & \textbf{ECE} $\downarrow$ & \textbf{Brier} $\downarrow$ & \textbf{ECE} $\downarrow$ & \textbf{Brier} $\downarrow$ & \textbf{ECE} $\downarrow$ & \textbf{Brier} $\downarrow$ & \textbf{ECE} $\downarrow$ & \textbf{Brier} $\downarrow$ & \textbf{ECE} $\downarrow$ & \textbf{Brier} $\downarrow$ \\
\midrule
DeepSeek-R1-7B        & Logit (Geom.)   & 0.145 & 0.233 & 0.012 & 0.214 & 0.013 & 0.214 & 0.007 & 0.214 & \textbf{0.109} & \textbf{0.230} \\
                      & Logit (P-true)  & 0.647 & 0.621 & 0.027 & 0.216 & 0.027 & 0.216 & 0.033 & 0.215 & 0.208 & 0.251 \\
                      & Verbal          & 0.428 & 0.438 & 0.033 & 0.212 & 0.019 & 0.216 & 0.042 & 0.219 & 0.191 & 0.252 \\
\addlinespace
Qwen2.5-7B            & Logit (Geom.)   & 0.076 & 0.260 & 0.012 & 0.249 & 0.012 & 0.249 & 0.019 & 0.251 & 0.058 & 0.251 \\
                      & Logit (P-true)  & 0.505 & 0.491 & 0.009 & 0.249 & 0.016 & 0.249 & 0.022 & 0.252 & 0.070 & 0.251 \\
                      & Verbal          & 0.394 & 0.404 & 0.074 & 0.247 & 0.073 & 0.252 & 0.074 & 0.252 & 0.067 & 0.251 \\
\addlinespace
DeepSeek-V3.2         & Verbal          & 0.290 & 0.351 & 0.044 & 0.249 & 0.041 & 0.250 & 0.041 & 0.250 & 0.054 & 0.251 \\
\addlinespace
Gemma-2-9B            & Logit (Geom.)   & 0.105 & 0.256 & 0.014 & 0.240 & 0.015 & 0.240 & 0.018 & 0.240 & 0.113 & 0.249 \\
                      & Logit (P-true)  & 0.459 & 0.438 & 0.031 & 0.235 & 0.025 & 0.235 & 0.019 & 0.234 & 0.123 & 0.250 \\
                      & Verbal          & 0.400 & 0.404 & 0.004 & 0.227 & 0.029 & 0.225 & 0.025 & 0.224 & 0.154 & 0.245 \\
\addlinespace
GPT-5-mini            & Verbal          & 0.316 & 0.353 & 0.060 & 0.249 & 0.042 & 0.251 & 0.047 & 0.251 & 0.044 & 0.252 \\
GPT-5-nano            & Verbal          & 0.236 & 0.327 & 0.057 & 0.249 & 0.057 & 0.249 & 0.057 & 0.248 & 0.129 & 0.308 \\
\addlinespace
Ministral-8B          & Logit (Geom.)   & 0.122 & 0.250 & 0.009 & 0.231 & 0.006 & 0.231 & 0.018 & 0.233 & 0.114 & 0.245 \\
                      & Logit (P-true)  & 0.268 & 0.296 & 0.044 & 0.230 & 0.038 & 0.230 & 0.019 & 0.228 & 0.147 & 0.250 \\
                      & Verbal          & 0.500 & 0.485 & 0.016 & 0.176 & 0.016 & 0.177 & 0.024 & 0.177 & 0.280 & 0.249 \\
\addlinespace
Qwen3.5-Flash         & Verbal          & 0.340 & 0.378 & 0.008 & 0.244 & 0.008 & 0.247 & 0.008 & 0.247 & 0.047 & 0.251 \\
\bottomrule
\end{tabular}%
}
\end{table}

\begin{table}[htbp]
\centering
\caption{Calibration performance on Formality dataset.}
\label{tab:formality}
\resizebox{\textwidth}{!}{%
\small
\begin{tabular}{@{}ll*{5}{cc}@{}}
\toprule
\textbf{Model} & \textbf{Method} & \multicolumn{2}{c}{\textbf{Original}} & \multicolumn{2}{c}{\textbf{Platt}} & \multicolumn{2}{c}{\textbf{Beta}} & \multicolumn{2}{c}{\textbf{Isotonic}} & \multicolumn{2}{c}{\textbf{Temperature}} \\
\cmidrule(lr){3-4} \cmidrule(lr){5-6} \cmidrule(lr){7-8} \cmidrule(lr){9-10} \cmidrule(l){11-12}
 & & \textbf{ECE} $\downarrow$ & \textbf{Brier} $\downarrow$ & \textbf{ECE} $\downarrow$ & \textbf{Brier} $\downarrow$ & \textbf{ECE} $\downarrow$ & \textbf{Brier} $\downarrow$ & \textbf{ECE} $\downarrow$ & \textbf{Brier} $\downarrow$ & \textbf{ECE} $\downarrow$ & \textbf{Brier} $\downarrow$ \\
\midrule
DeepSeek-R1-7B        & Logit (Geom.)   & 0.138 & 0.240 & 0.037 & 0.220 & 0.036 & 0.220 & 0.033 & 0.220 & \textbf{0.112} & \textbf{0.241} \\
                      & Logit (P-true)  & 0.639 & 0.611 & 0.033 & 0.220 & 0.033 & 0.220 & 0.038 & 0.221 & 0.196 & 0.251 \\
                      & Verbal          & 0.373 & 0.365 & 0.023 & 0.194 & 0.020 & 0.194 & 0.019 & 0.194 & 0.234 & 0.249 \\
\addlinespace
Qwen2.5-7B            & Logit (Geom.)   & 0.183 & 0.255 & 0.056 & 0.224 & 0.056 & 0.224 & 0.064 & 0.225 & 0.186 & 0.250 \\
                      & Logit (P-true)  & 0.646 & 0.614 & 0.055 & 0.224 & 0.056 & 0.224 & 0.056 & 0.222 & 0.193 & 0.252 \\
                      & Verbal          & 0.457 & 0.441 & 0.027 & 0.218 & 0.027 & 0.218 & 0.027 & 0.218 & 0.194 & 0.251 \\
\addlinespace
DeepSeek-V3.2         & Verbal          & 0.380 & 0.395 & 0.011 & 0.228 & 0.011 & 0.228 & 0.011 & 0.228 & 0.157 & 0.250 \\
\addlinespace
Gemma-2-9B            & Logit (Geom.)   & 0.122 & 0.253 & 0.008 & 0.235 & 0.008 & 0.234 & 0.013 & 0.235 & 0.129 & 0.249 \\
                      & Logit (P-true)  & 0.509 & 0.501 & 0.008 & 0.234 & 0.008 & 0.234 & 0.014 & 0.232 & 0.134 & 0.251 \\
                      & Verbal          & 0.431 & 0.428 & 0.024 & 0.230 & 0.024 & 0.230 & 0.024 & 0.230 & 0.142 & 0.249 \\
\addlinespace
GPT-5-mini            & Verbal          & 0.431 & 0.418 & 0.029 & 0.240 & 0.028 & 0.240 & 0.029 & 0.240 & 0.112 & 0.251 \\
GPT-5-nano            & Verbal          & 0.384 & 0.377 & 0.010 & 0.220 & 0.011 & 0.219 & 0.011 & 0.219 & 0.153 & 0.245 \\
\addlinespace
Ministral-8B          & Logit (Geom.)   & 0.156 & 0.244 & 0.015 & 0.215 & 0.024 & 0.215 & 0.012 & 0.213 & 0.178 & 0.243 \\
                      & Logit (P-true)  & 0.313 & 0.314 & 0.049 & 0.218 & 0.051 & 0.218 & 0.043 & 0.222 & 0.194 & 0.251 \\
                      & Verbal          & 0.457 & 0.461 & 0.034 & 0.199 & 0.034 & 0.199 & 0.033 & 0.199 & 0.227 & 0.250 \\
\addlinespace
Qwen3.5-Flash         & Verbal          & 0.451 & 0.448 & 0.013 & 0.239 & 0.016 & 0.239 & 0.011 & 0.239 & 0.106 & 0.251 \\
\bottomrule
\end{tabular}%
}
\end{table}

\begin{table}[htbp]
\centering
\caption{Calibration performance on Politeness dataset.}
\label{tab:politeness}
\resizebox{\textwidth}{!}{%
\small
\begin{tabular}{@{}ll*{5}{cc}@{}}
\toprule
\textbf{Model} & \textbf{Method} & \multicolumn{2}{c}{\textbf{Original}} & \multicolumn{2}{c}{\textbf{Platt}} & \multicolumn{2}{c}{\textbf{Beta}} & \multicolumn{2}{c}{\textbf{Isotonic}} & \multicolumn{2}{c}{\textbf{Temperature}} \\
\cmidrule(lr){3-4} \cmidrule(lr){5-6} \cmidrule(lr){7-8} \cmidrule(lr){9-10} \cmidrule(l){11-12}
 & & \textbf{ECE} $\downarrow$ & \textbf{Brier} $\downarrow$ & \textbf{ECE} $\downarrow$ & \textbf{Brier} $\downarrow$ & \textbf{ECE} $\downarrow$ & \textbf{Brier} $\downarrow$ & \textbf{ECE} $\downarrow$ & \textbf{Brier} $\downarrow$ & \textbf{ECE} $\downarrow$ & \textbf{Brier} $\downarrow$ \\
\midrule
DeepSeek-R1-7B        & Logit (Geom.)   & 0.209 & 0.238 & 0.007 & 0.190 & 0.006 & 0.190 & 0.028 & 0.191 & \textbf{0.177} & \textbf{0.241} \\
                      & Logit (P-true)  & 0.689 & 0.670 & 0.006 & 0.188 & 0.006 & 0.189 & 0.020 & 0.189 & 0.259 & 0.252 \\
                      & Verbal          & 0.377 & 0.374 & 0.011 & 0.150 & 0.011 & 0.150 & 0.011 & 0.150 & 0.307 & 0.248 \\
\addlinespace
Qwen2.5-7B            & Logit (Geom.)   & 0.119 & 0.267 & 0.012 & 0.246 & 0.016 & 0.247 & 0.037 & 0.248 & 0.088 & 0.250 \\
                      & Logit (P-true)  & 0.529 & 0.518 & 0.018 & 0.249 & 0.015 & 0.249 & 0.037 & 0.252 & 0.093 & 0.251 \\
                      & Verbal          & 0.382 & 0.380 & 0.036 & 0.237 & 0.029 & 0.238 & 0.035 & 0.234 & 0.135 & 0.251 \\
\addlinespace
DeepSeek-V3.2         & Verbal          & 0.272 & 0.333 & 0.032 & 0.243 & 0.031 & 0.243 & 0.034 & 0.244 & 0.093 & 0.250 \\
\addlinespace
Gemma-2-9B            & Logit (Geom.)   & 0.266 & 0.263 & 0.003 & 0.182 & 0.004 & 0.182 & 0.013 & 0.184 & 0.267 & 0.250 \\
                      & Logit (P-true)  & 0.637 & 0.604 & 0.004 & 0.181 & 0.004 & 0.182 & 0.004 & 0.183 & 0.270 & 0.252 \\
                      & Verbal          & 0.543 & 0.498 & 0.033 & 0.154 & 0.034 & 0.154 & 0.038 & 0.159 & 0.308 & 0.249 \\
\addlinespace
GPT-5-mini            & Verbal          & 0.690 & 0.592 & 0.011 & 0.105 & 0.004 & 0.109 & 0.004 & 0.109 & 0.382 & 0.251 \\
GPT-5-nano            & Verbal          & 0.365 & 0.391 & 0.013 & 0.237 & 0.013 & 0.237 & 0.032 & 0.237 & 0.097 & 0.250 \\
\addlinespace
Ministral-8B          & Logit (Geom.)   & 0.244 & 0.257 & 0.049 & 0.187 & 0.049 & 0.187 & 0.062 & 0.188 & 0.261 & 0.249 \\
                      & Logit (P-true)  & 0.414 & 0.356 & 0.050 & 0.188 & 0.051 & 0.188 & 0.065 & 0.190 & 0.267 & 0.251 \\
                      & Verbal          & 0.516 & 0.483 & 0.015 & 0.129 & 0.015 & 0.129 & 0.016 & 0.130 & 0.350 & 0.251 \\
\addlinespace
Qwen3.5-Flash         & Verbal          & 0.487 & 0.467 & 0.004 & 0.209 & 0.005 & 0.210 & 0.005 & 0.210 & 0.196 & 0.249 \\
\bottomrule
\end{tabular}%
}
\end{table}

% \section{Technical appendices and supplementary material}
% Technical appendices with additional results, figures, graphs, and proofs may be submitted with the paper submission before the full submission deadline (see above). You can upload a ZIP file for videos or code, but do not upload a separate PDF file for the appendix. There is no page limit for the technical appendices. 

% Note: Think of the appendix as ``optional reading'' for reviewers. The paper must be able to stand alone without the appendix, for example, adding critical experiments that support the main claims to an appendix is inappropriate. 

%%%%%%%%%%%%%%%%%%%%%%%%%%%%%%%%%%%%%%%%%%%%%%%%%%%%%%%%%%%%

\newpage
% \input{checklist.tex}

% \section{Prompt Template}

% \section{Detailed Results per Dataset}
% \label{app:full_results}
% % 放全量表格

% \section{Extended Reliability Diagrams}
% \label{app:diagrams}
% % 放各种图

% \section{Computational Resources and Hyperparameters}
% \label{app:reproducibility}
% % 放算力和训练细节

\end{document}